%% file: LSK3DNet.tex
\documentclass[10pt,twocolumn,letterpaper]{article}

\usepackage[pagenumbers]{cvpr} 

\input{preamble}

\definecolor{cvprblue}{rgb}{0.21,0.49,0.74}
\usepackage[pagebackref,breaklinks,colorlinks,citecolor=cvprblue]{hyperref}

\usepackage[capitalize]{cleveref}
\crefname{section}{Sec.}{Secs.}
\Crefname{section}{Section}{Sections}
\Crefname{table}{Table}{Tables}
\crefname{table}{Tab.}{Tabs.}

\def\paperID{3635} 
\def\confName{CVPR}
\def\confYear{2024}

\title{\!\!\!\!\!LSK3DNet: Towards Effective and Efficient 3D Perception with Large Sparse Kernels\!\!\!\!\!}

\author{
Tuo Feng$^1$, Wenguan Wang$^2$, Fan Ma$^2$, Yi Yang$^2$
\\
{\small {$^1$}  ReLER, AAII, University of Technology Sydney \hspace{0pt}} 
{\small{$^2$}  ReLER, CCAI, Zhejiang University \hspace{0pt}} 
\\
{\small \url{https://github.com/FengZicai/LSK3DNet}}
}

\begin{document}
\maketitle
\begin{abstract}
Autonomous systems need to process large-scale, sparse, and irregular point clouds with limited compute resources. Consequently, it is essential to develop LiDAR perception methods that are both efficient and effective.$_{\!}$ Although naive- ly enlarging 3D kernel size can enhance performance, it will also lead to a cubically-increasing overhead. Therefore, it is crucial to develop streamlined 3D large kernel designs that eliminate redundant weights and work effectively with larger kernels. In this paper, we propose an efficient and effective {Large Sparse Kernel 3D Neural Network} ({LSK3DNet}) that leverages dynamic pruning to amplify the 3D kernel size. Our method comprises two core components: \textit{Spatial-wise Dynamic Sparsity} ({SDS}) and \textit{Channel-wise Weight Selection} ({CWS}). SDS dynamically prunes and regrows volumetric weights from the beginning to learn a large sparse 3D kernel. It not only boosts performance but also significantly reduces model size and computational cost.$_{\!}$ Moreover, CWS selects the most important channels for 3D convolution during training and subsequently prunes the redundant channels to accelerate inference for 3D vision tasks. We demonstrate the effectiveness of LSK3DNet on three benchmark datasets and five tracks compared with classical models and large kernel designs. Notably, LSK3DNet achieves the state-of-the-art performance on SemanticKITTI (\ie, 75.6\% on single-scan and 63.4\% on multi-scan), with roughly 40\% model size reduction and 60\% computing operations reduction compared to the naive large 3D kernel model. 
\end{abstract}

\vspace{-10pt}
\section{Introduction}
\vspace{-3pt}
\label{sec:introduction}
Autonomous systems, such as intelligent robots and self-driving cars, are equipped with 3D LiDAR sensors and data processing platforms. The LiDAR sensor generates point clouds, which serve as the input for the processing platform. The platform performs perception tasks such as semantic segmentation~\cite{zhou2021adaptive, feng2023clustering, yan20222dpass} and object detection~\cite{shi2020pv, deng2021voxel, yin2024instance}, providing a foundation for motion planning and decision-making~\cite{feng2024interpretable3D} in autonomous systems. On the one hand, the processing platform is required to handle large-scale, sparse, and irregular point clouds. On the other hand, the platform’s computing resources are limited. Therefore, it is crucial to develop effective and efficient LiDAR perception methods. Point-based methods~\cite{zhou2021adaptive,zhao2021point,lai2022stratified} rely on computationally expensive, memory inefﬁcient, and time-consuming point sampling strategy~\cite{hu2019randla}. In contrast, sparse convolution~\cite{graham20183d,choy20194d,yin2022semi,yin2022proposalcontrast} has been widely adopted for processing large-scale point clouds up to $160\!\times\!160\!\times\!20$ meters~\cite{choy20194d, graham20183d,zhu2021cylindrical,tang2020searching, deng2021voxel, yin2021center}.
\begin{figure}[t]
  \centering
  \vspace{-6pt}
      \includegraphics[width=0.97 \linewidth]{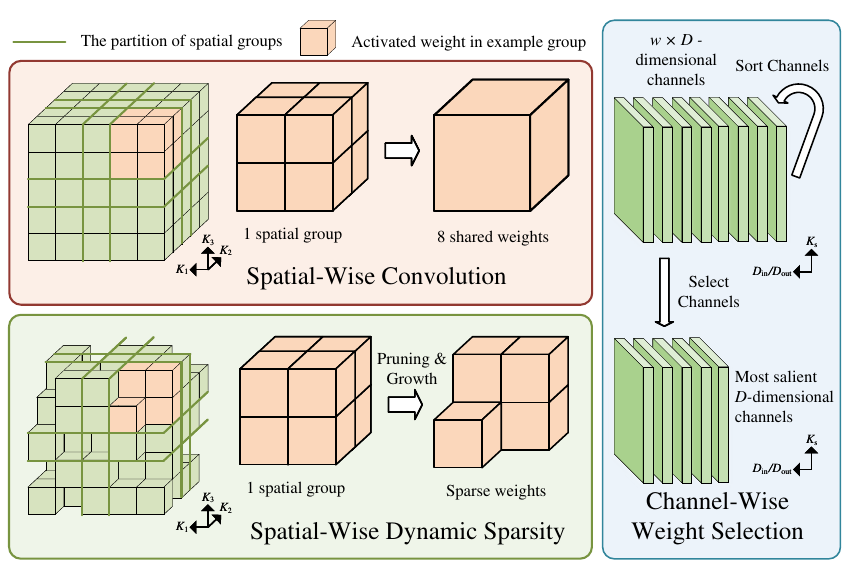}
      \put(-91,78){\scalebox{.66}{\cite{chen2022scaling}}}
\vspace{-11pt}
\captionsetup{font=small}
\caption{\small Illustrations on \textit{Spatial-wise Group Convolution}~\cite{chen2022scaling}, SDS, and CWS. The spatial dimensions $K_s$ (\ie, $K_1$, $K_2$, $K_3$) and channel dimensions ($D_{in}$/$D_{out}$) are shown. \textit{Spatial-wise Group Convolution} shares the weights within each spatial group during training, leading to redundant model weights. In contrast, SDS removes non-salient weights and redundancies that are not sensitive to the input in each group, while CWS eliminates redundant weights in a channel-wise manner (Sec.\!~\ref{sec:introduction}).}
\label{fig:conv}
\vspace{-21pt}
\end{figure}
Nevertheless, the application of such a technique in autonomous systems has also been hindered by limited computational resources~\cite{tang2020searching}. PV-KD~\cite{hou2022point} attempts to reduce the model size with knowledge distillation, but its performance is bounded by its teacher model~\cite{zhu2021cylindrical}. In opposition, large kernels can enhance performance through the advantages of large receptive fields~\cite{ding2022scaling,liu2022more, chen2022scaling}. However, naive 3D large kernels will significantly increase computational burdens. LargeKernel3D~\cite{chen2022scaling} explores large 3D kernels with \textit{Spatial-wise Group Convolution} to circumvent optimization and efficiency problems caused by naive 3D large kernels. However, LargeKernel3D has a lot of redundant model weights during training as it shares the weights in each spatial group (See Fig.\!~\ref{fig:conv}). Moreover, the performance of LargeKernel3D~\cite{chen2022scaling} drops when scaling up the kernel size over $7\!\times\!7\!\times\!7$.

\begin{figure}[t]
  \centering
  \vspace{-6pt}
      \includegraphics[width=0.99 \linewidth]{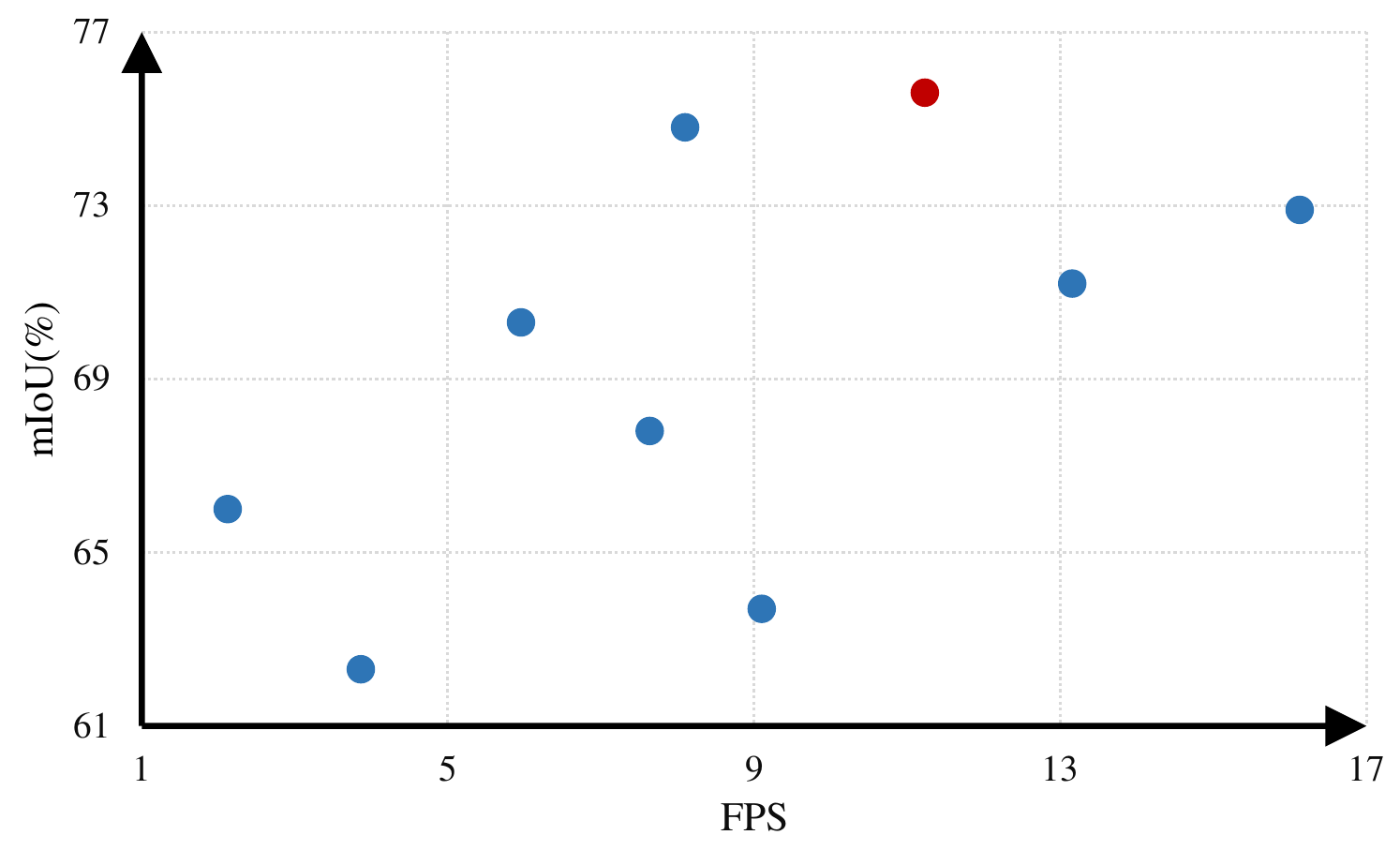}
    \put(-193,55){\scalebox{.70}{$\texttt{JS3C-Net}$\cite{yan2021sparse}}}
    \put(-191,34.5){\scalebox{.70}{$\texttt{SPVNAS}$\cite{tang2020searching}}}
    \put(-126,44.5){\scalebox{.70}{$\texttt{MinkUNet}$\cite{choy20194d}}}
    \put(-122.5,68.5){\scalebox{.70}{$\texttt{Cylinder3D}$\cite{zhu2021cylindrical}}}
    \put(-144,86.5){\scalebox{.70}{$\texttt{RPVNet}$\cite{xu2021rpvnet}}}
    \put(-148,113){\scalebox{.70}{$\texttt{SphereFormer}$\cite{lai2023spherical}}}
    \put(-76,125){\scalebox{.70}{$\textbf{\texttt{LSK3DNet}}$}}
    \put(-69.5,87){\scalebox{.70}{$\texttt{PV-KD}$\cite{hou2022point}}}
    \put(-55,105.5){\scalebox{.70}{$\texttt{2DPASS}$\cite{yan20222dpass}}}
    \put(-101,4){\scalebox{.65}{($\uparrow$)}}
\vspace{-10pt}
\captionsetup{font=small}
\caption{\small $_{\!}$Performance$_{\!}$ (mIoU)$_{\!}$ \textit{vs}.$_{\!}$ Inference$_{\!}$ Speed$_{\!}$ (FPS)$_{\!}$ on$_{\!}$ SemanticKITTI~\cite{behley2019semantickitti} single-scan challenge (Sec.\!~\ref{sec:introduction}).}
\label{fig:speed}
\vspace{-16pt}
\end{figure}
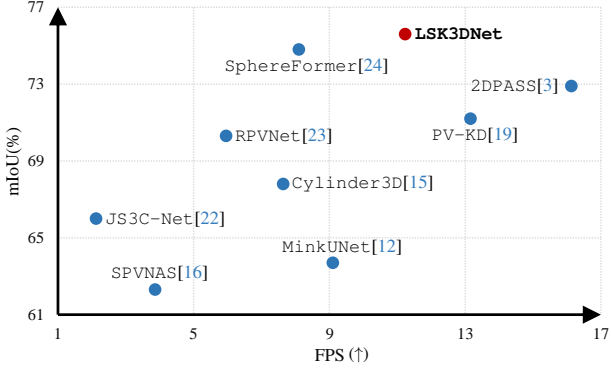


In this study, we propose a Large Sparse Kernel 3D Neural Network (LSK3DNet), which is an efficient and effective 3D perception model. LSK3DNet incorporates two key components, namely \textit{Spatial-wise Dynamic Sparsity} ({SDS}) and \textit{Channel-wise Weight Selection} ({CWS}). These components enhance performance on perception tasks and overcome the challenges of high computational costs. SDS enlarges 3D kernels and reduces model parameters by learning sparse 3D kernels from scratch. CWS increases the model width and learns channel-wise importance during training, resulting in accelerated inference by pruning redundant channels. As shown in Fig.\!~\ref{fig:conv}, SDS can prune 3D kernels to remove redundant weights, while CWS can eliminate redundant channels in an online manner. Our SDS and CWS complement each other following a guideline of \textit{``using spatial sparse groups, expanding width without more parameters"}, in contrast to \textit{``using sparse groups, expanding more"}~\cite{liu2022more}.

Specifically, SDS implements a remove-and-regrow process within each spatial group and preserves the sparsity in each group during dynamic training. This stands in stark contrast to the 2D approach~\cite{liu2022more}, as it achieves dynamic sparsity in depth-wise groups. SDS allows to scale up the receptive fields with large kernel sizes, easily reaching or surpassing $9\!\times\!9\!\times\!9$, thereby achieving higher performance. In addition, CWS selectively identifies salient channels during training. It speeds up inference on the 3D vision tasks by pruning non-salient channel parameters. In this way, LSK3DNet can benefit from expanded width to achieve enhanced performance, but maintain the original model size during inference. Reducing the models’ complexity (\eg, model size) is a big benefit and a straightforward way to make them deployable on resource-constrained devices~\cite{kamath2023deep}. We conduct experiments on three benchmark datasets and five tracks.$_{\!}$ Our method achieves better performance compared to state-of-the-art methods~\cite{lai2023spherical,xu2021rpvnet,zhu2021cylindrical} on SemanticKITTI~\cite{behley2019semantickitti} but with faster inference speed (Fig.\!~\ref{fig:speed}). Moreover, LSK3DNet outperforms the prior 3D large kernel method of LargeKernel3D~\cite{chen2022scaling} on ScanNet v2~\cite{dai2017scannet} and KITTI~\cite{geiger2013vision}.

In a nutshell, our \textbf{contributions} are as follows:
\begin{itemize}[leftmargin=*]
    \setlength{\itemsep}{4pt}
    \setlength{\parsep}{-2pt}
    \setlength{\parskip}{-0pt}
    \setlength{\leftmargin}{-10pt}
    \item We propose LSK3DNet to enhance performance in perception tasks and mitigate high computational costs, surpassing state-of-the-art models while reducing model size by 40\% and computing operations by 60\%.
    \item We propose SDS to scale up 3D kernels. It learns large sparse kernels by dynamically pruning and regrowing weights from scratch, thereby reducing model size and computational operations.
    \item We develop CWS to improve performance by expanding the width. CWS assesses the importance of channels during training and speeds up inference by pruning redundant channel-wise parameters. 
\end{itemize}

\vspace{-3pt}
\section{Related Work}%
\vspace{-1pt}

\noindent\textbf{Large-Kernel 3D Models.}~In 3D vision, research on 3D large kernel is very limited. LargeKernel3D~\cite{chen2022scaling} has demonstrated that sizeable kernels can be successfully employed and bring positive results for 3D networks.$_{\!}$ \textit{Spatial-wise Group Convolution}~\cite{chen2022scaling} enables to achieve a kernel size of $7\!\times\!7\!\times\!7$. However, it shares the weights within each spatial group during training, leading to redundant model weights (See Fig.\!~\ref{fig:conv}). Moreover, the performance of LargeKernel3D drops when scaling up the kernel size over $7\!\times\!7\!\times\!7$.

\noindent\textbf{Large-Kernel$_{\!}$ 2D$_{\!}$ CNN$_{\!}$ Models.}~In the 2010s, various large-kernel settings were investigated. LR-Net~\cite{hu2019local}, Inceptions~\cite{szegedy2016rethinking, szegedy2017inception}, and GCNs~\cite{peng2017large} explored 2D large kernels of $7\times7$, $11\times11$, and $15\times15$ respectively. Due to the widespread adoption~\cite{wang2022looking} of VGG~\cite{simonyan2014very}, research into large kernels was largely overlooked in favour of multiple smaller kernels ($1\!\times\!1$ or $3\!\times\!3$) to obtain a larger receptive field~\cite{he2016deep,howard2017mobilenets,huang2017densely,xie2017aggregated} during the past decade. Recently, certain studies have reintroduced large kernels in CNNs.$_{\!}$ RepLKNet~\cite{ding2022scaling} examines the effects of large kernels in CNNs, and for the first time it is able to increase kernel size to $31\!\times\!31$. It achieves comparable results to those of Swin Transformer~\cite{liu2021swin}.$_{\!}$ Following \textit{``use sparse groups, expand more''}, SLaK~\cite{liu2022more} has achieved an impressive kernel size of $51\!\times\!51$.$_{\!}$ SLaK has to strike a trade-off between sparsity and model width since increased model width leads to an increase in model size, a phenomenon that becomes even more pronounced in 3D vision.$_{\!}$ However, CWS effectively decouples the two objectives of improved performance and reduced model size, enabling expanded width without increasing the model size.

\noindent\textbf{3D Backbones.}~Generally speaking, there are three mainstreams of 3D backbones: \textit{i)} \textit{Point-based} methods~\cite{qi2017pointnet,qi2017pointnet++,thomas2019kpconv,xu2021paconv,fan2021scf,ran2021learning,xiang2021walk,liu2019relation,xu2020grid,zhou2021adaptive,wang2019graph,lian2019dense,qian2021pu,zhao2021point,lai2022stratified,fan2021deep,fan2022point,fan2020pstnet,fan2021point} approximate a permutation-invariant set function, and directly learn features from raw point clouds. Researchers have observed an unsatisfactory outcome for large-scale environments~\cite{hu2021towards}. \textit{ii)} \textit{Projection-based} methods transform unstructured points into regular 2D images and thus employ CNNs~\cite{tatarchenko2018tangent,wu2018squeezeseg,wu2019squeezesegv2,cortinhal2020salsanext,xu2020squeezesegv3,zhang2020polarnet}. However, the projection could leave out critical geometric details and cause unavoidable information loss.$_{\!}$ \textit{iii)} \textit{Voxel-based} methods have used sparse convolution~\cite{choy20194d, graham20183d} to design more powerful networks for various 3D tasks~\cite{zhu2021cylindrical,yan2018second}.$_{\!}$ Sparse convolution facilitates convolution operations only on non-void voxels.$_{\!}$ Recently, \textit{fusion-based} methods~\cite{tang2020searching, xu2021rpvnet, yan20222dpass} have become upsurge. The point-based high-resolution branch mitigates the performance degradation, which is caused by aggressive downsampling~\cite{tang2020searching,chen2022focal} (\ie, regular sparse convolution) of the voxel branch.

\noindent\textbf{Dynamic$_{\!}$ Sparse$_{\!}$ Training (DST).}~DST can train$_{\!}$ sparse$_{\!}$ neural$_{\!}$ networks$_{\!}$ from$_{\!}$ scratch,$_{\!}$ resulting$_{\!}$ in$_{\!}$ both$_{\!}$ a$_{\!}$ speedy$_{\!}$ training$_{\!}$ and$_{\!}$ prediction$_{\!}$ procedure.$_{\!}$ During$_{\!}$ training,$_{\!}$ DST~\cite{bellec2017deep,mocanu2018scalable,dettmers2019sparse,liu2021sparse,evci2020rigging,mostafa2019parameter,jayakumar2020top,chen2021chasing,liu2021selfish}$_{\!}$ alters$_{\!}$ the$_{\!}$ location$_{\!}$ of$_{\!}$ non-zero$_{\!}$ weights$_{\!}$ with$_{\!}$ per pre-defined$_{\!}$ rules,$_{\!}$ thereby$_{\!}$ creating$_{\!}$ a$_{\!}$ sparse$_{\!}$ representation$_{\!}$ and$_{\!}$ cutting$_{\!}$ down$_{\!}$ the$_{\!}$ number$_{\!}$ of$_{\!}$ calculations.$_{\!}$ This$_{\!}$ paradigm$_{\!}$ commences$_{\!}$ without$_{\!}$ prior$_{\!}$ knowledge$_{\!}$ and$_{\!}$ simultaneously$_{\!}$ refines$_{\!}$ the$_{\!}$ non-zero$_{\!}$ locations$_{\!}$ and$_{\!}$ weights.$_{\!}$ The$_{\!}$ attractive$_{\!}$ aspect$_{\!}$ of$_{\!}$ DST$_{\!}$ is$_{\!}$ that$_{\!}$ it$_{\!}$ is$_{\!}$ sparse$_{\!}$ from$_{\!}$ the$_{\!}$ beginning,$_{\!}$ resulting$_{\!}$ in$_{\!}$ lower$_{\!}$ FLOPs$_{\!}$ and$_{\!}$ memory$_{\!}$ consumption$_{\!}$ for$_{\!}$ training$_{\!}$ and$_{\!}$ inference$_{\!}$ compared$_{\!}$ to$_{\!}$ a$_{\!}$ dense$_{\!}$ model.$_{\!}$ Generally,$_{\!}$ the$_{\!}$ pruning$_{\!}$ process~\cite{han2016dsd}$_{\!}$ can$_{\!}$ be$_{\!}$ completed$_{\!}$ either$_{\!}$ by$_{\!}$ threshold-based$_{\!}$ pruning~\cite{bellec2017deep,mocanu2017network,mocanu2018scalable,mostafa2019parameter}$_{\!}$ or$_{\!}$ by$_{\!}$ magnitude-based$_{\!}$ pruning~\cite{dettmers2019sparse,dai2019nest}.$_{\!}$ In$_{\!}$ addition,$_{\!}$ new$_{\!}$ weights$_{\!}$ are$_{\!}$ regrown$_{\!}$ with$_{\!}$ randomness$_{\!}$ growing~\cite{mocanu2017network,mocanu2018scalable,mostafa2019parameter},$_{\!}$ momentum$_{\!}$ growing~\cite{dettmers2019sparse},$_{\!}$ and$_{\!}$ gradient-based$_{\!}$ growing~\cite{evci2020rigging,dai2019nest,dai2019grow,liu2021sparse}.$_{\!}$ Here, we efficiently integrate dynamic sparsity with 3D kernels in large kernel 3D networks.$_{\!}$ This design satisfies the anticipation for a larger parameter space, as a wide-ranging exploration of the parameter space is of significance to DST~\cite{jayakumar2020top, raihan2020sparse, liu2021selfish}.

\vspace{-3pt}
\section{Methodology}
\vspace{-1pt}

\label{sec:method}
In this section, we first formulate the problem in Sec.\!~\ref{sec:formulation}, then move on to a concise overview of submanifold sparse convolution in Sec.\!~\ref{sec:Submanifold}. Afterwards, the details of \textit{Spatial-wise Dynamic Sparsity} ({SDS}) and \textit{Channel-wise Weight Selection} ({CWS}) are described in Sec.\!~\ref{sec:Sparsity} and\!~\ref{sec:Reconstitution}. Lastly, network architecture is outlined in Sec.\!~\ref{sec:Objective}.

\subsection{Problem Formulation}
\vspace{-1pt}
\label{sec:formulation}
For 3D perception tasks, suppose that the input is a set of unordered point $\mathcal{X}\!=\!\left\{x_{i} \in \mathbb{R}^{D_{in}}: i=1, 2, \ldots, N\right\}$, and $D_{in}$ is the dimensionality of the input features.$_{\!}$ $\mathcal{L}$ is a set of object/point labels, which varies according to different datasets. For segmentation task, {LSK3DNet} assigns each point $x_{i}$ with a predicted label $y_{i}$, resulting in a point-wise prediction set $\mathcal{Y}\!=\left\{y_{i}: i=1, 2, \ldots, N\right\}$.$_{\!}$ In voxel-based methods, point clouds are converted to voxel representations.$_{\!}$ We denote the input sparse tensor with $X_{s}^{in}=[C, F]$.$_{\!}$ Coordinate matrix $C\!\in\!\mathbb{N}^{N \times 3\!}_+$ includes 3-dimensional discrete coordinates, and feature matrix $F\!\in\!\mathbb{R}^{N \times D_{in}}$ has $D_{in}$-dimensional features.

\begin{figure*}[t]
  \centering
  \vspace{-6pt}
      \includegraphics[width=0.98 \linewidth]{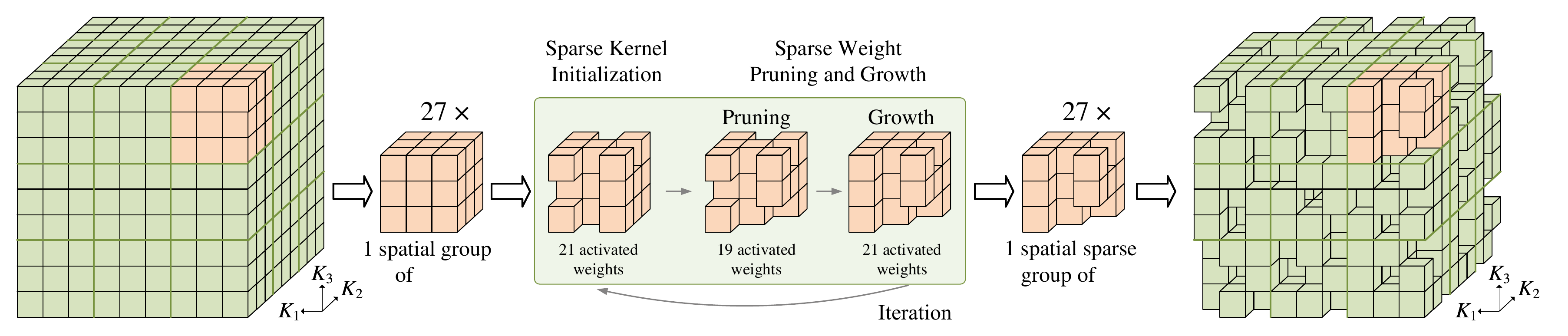}
      \put(-356,16.5){\scalebox{.65}{$\bm{W}_{\!D}$}}
      \put(-144.5,16.5){\scalebox{.65}{$\bm{W}_{\!S}$}}
\vspace{-12pt}
\captionsetup{font=small}
\caption{\small \textbf{Spatial-wise Dynamic Sparsity.} The utilization of SDS enables us to create and train sparse kernel 3D neural networks from the beginning. The sparse weights in each spatial group are firstly initialized by \textit{Sparse Kernel Initialization} (Eq.$_{\!}$~(\ref{eq:Initialization})), and then regularly altered by discarding the least significant connections and introducing new ones (Eq.$_{\!}$~(\ref{eq:pruning})). The sparse kernels are steadily improved by this dynamic process, which leads to a more thorough collection of local features. Note that different spatial sparse group has different sparse distribution. Here we take a sparsity of 22\% for example (Sec.\!~\ref{sec:Sparsity} and Alg.\!~\ref{alg:algorithm1}).}
\label{fig:SDS}
\vspace{-14pt}
\end{figure*}

\subsection{Review of Submanifold Sparse Convolution}
\vspace{-1pt}
\label{sec:Submanifold} 
Regular sparse convolution has widespread use in U-Net type 3D networks~\cite{graham20183d,choy20194d}.$_{\!}$ However, it significantly reduces the sparsity level of point data~\cite{chen2022focal}, obfuscates feature distinctions~\cite{chen2022focal}, and leads to low-resolution and indistinguishable small objects~\cite{tang2020searching}.$_{\!}$ In our LSK3DNet, we discard regular sparse convolution and exclusively employ submanifold sparse convolution for feature extraction.

In a $3$-dimensional space $C\!\in\!\mathbb{N}^{N \times 3\!}_+$, the $D_{in}$-dimensional input feature at $c\!\in\!C$ can be denoted as $x^{in}_c\!\in\!\mathbb{R}^{D_{in}}$, and the 3D kernel is represented by $\bm{W} \!\in\!\mathbb{R}^{K^3\!\times\!D_\text{out}\!\times\!D_\text{in}}$.$_{\!}$ We divide the kernel into $K^3$ spatial weights, each with a size of $D_{out}\!\times\!D_{in}$, and denote the divided weights as $\bm{W}_{i}$. The submanifold sparse convolution is formulated as follows:

\vspace{-2pt}
\begin{equation}\small
x^{{out}}_c = \sum_{i \in \mathcal{V}^D(c, C)} \bm{W}_i \thinspace x^{in}_{c + i}, \quad c \in C,
\label{eq:submanifold_sparse_convolution}
\vspace{-2pt}
\end{equation}
where $x_c^{out}$ represents the current voxel on which the submanifold sparse convolution is applied; $\bm{W}_{i}$ corresponds to $x_{c+i}^{in}$, which represents the $i$-th adjacent input voxel ($[C_{c+i}, F_{c+i}]$); $\mathcal{V}^D$ is a set of offsets that define the shape of a kernel, and $\mathcal{V}^D(c, C)\!=\!\{i | c + i\!\in\!C, i\!\in\!\mathcal{V}^D\!\}$ is a set of offsets from the current center $c$ that exist in $C$.$_{\!}$ It should be noted that the input coordinates and output coordinates are equivalent; in other words, they share the same $C$.$_{\!}$ Due to this restriction, insufficient information flow exists while differentiating the distinct spatial characteristics~\cite{chen2022focal}. Large receptive fields could be a potential solution to this problem.

\subsection{Spatial-wise Dynamic Sparsity ({SDS})}
\vspace{-1pt}
\label{sec:Sparsity}
{SDS} incorporates two essential components: \textit{Sparse Kernel Initialization} and \textit{Sparse Weight Pruning and Growth}.$_{\!}$ SDS is applied to the 3D large kernel layers, while other layers are kept dense.$_{\!}$ When initialized with \textit{Sparse Kernel Initialization}, all spatial groups of sparse tensors have the same level of sparsity $s$, where sparsity refers to the fraction of zeros in sparse kernels.$_{\!}$ We adjust the sparse position within spatial groups of 3D large kernel layers with an adaptation frequency $f_a$.$_{\!}$ During training, at regular intervals determined by $f_a$, the adjustable weights in the kernels are modified through a two-phase procedure, which includes \textit{sparse weight pruning} and \textit{sparse weight growth}.$_{\!}$ We maintain a constant number of non-zero parameters (\ie, sparsity $s$) throughout the training process.

\noindent\textbf{Sparse Kernel Initialization.}~Prior research~\cite{lee2018snip} has explored the distribution of sparsity parameters across multiple layers.$_{\!}$ The proportion and adjustment of sparse weights are regulated by the layer-wise sparsity ratio, which determines how the weights are modified to seek more efficient sparse structures during training. However, this method is not suitable for \textit{Spatial-wise Group Convolution}, since 3D large kernels are divided into multiple spatial groups. 
To ensure that each group has non-zero weights, we execute Erd{\H{o}}s-R{\'e}nyi (ER) based sparsity ratios~\cite{mocanu2018scalable, evci2020rigging, liu2022the} in each spatial group. Dense weights $\bm{W}_{\!D}$ are firstly initialized in a standard way (\ie, kaiming uniform initialization~\cite{he2015delving}). The binary Mask $\bm{M}$ is then applied to these dense weights; $\bm{M}$ is generated using an ER-based method with a sparsity of $s$. The zero-weight number in $\bm{M}$ is scaled by \begin{footnotesize}$1\!-\!\frac{D_{in}\!+\!D_{out}\!+\!K_{1}^{g}\!+\!K_{2}^{g}\!+\!K_{3}^{g}}{D_{in}\!\times\!D_{out}\!\times\!K_{1}^{g}\!\times\!K_{2}^{g}\!\times\!K_{3}^{g}}$\end{footnotesize}, where \begin{footnotesize}$K_1^g$\end{footnotesize}, \begin{footnotesize}$K_2^g$\end{footnotesize}, and \begin{footnotesize}$K_3^g$\end{footnotesize} denote the spatial group sizes. Therefore, the sparse weights $\bm{W}_{\!S}$ are initialized as follows:
\vspace{-2pt} 
\begin{equation}\small
  \begin{gathered}\small\label{eq:Initialization}
    \bm{W}_{\!S} \leftarrow \bm{W}_{\!D} \odot \bm{M},
  \end{gathered}
\vspace{-2pt}
\end{equation}
where $\odot$ represents the Hadamard product.

To design 3D networks with sparse groups, following the recipe of \textit{``using spatial sparse groups, expanding width without more parameters"}, we have chosen \textit{Spatial-wise Group Convolution} rather than \textit{Depth-wise Group Convolution}. Our experiments have also shown that \textit{Depth-wise Group Convolution} is unsuitable for large 3D kernels. This aligns with the findings in LargeKernel3D~\cite{chen2022scaling}. This experiment is provided in the supplementary materials.

\noindent\textbf{Sparse$_{\!}$ Weight$_{\!}$ Pruning$_{\!}$ and$_{\!}$ Growth.}~Previous research~\cite{liu2022more} in 2D convolution has illustrated that dynamic sparsity can effectively enlarge kernel sizes and enhance scalability. But \textit{Depth-wise Group Convolution} is unsuitable for large 3D kernels~\cite{chen2022scaling}. Therefore, we adapt sparse weights in each spatial group dynamically to accommodate the data. Specifically, a certain rate of connections (\ie, prune rate $p$) with the lowest magnitude are eliminated, and then we generate an equal number of new random connections. The regrow weights are randomly placed at zero positions within each spatial group. This is achieved by modifying the mask $\bm{M}$ (See Alg.\!~\ref{alg:algorithm1}). The positions of the eliminated connections and the newly generated connections are denoted as $\bm{E}$ and $\bm{G}$, respectively. The formula for pruning and regrowth can be expressed as follows:
\vspace{-2pt}
\begin{equation}\small
  \begin{gathered}\small\label{eq:pruning}
    \bm{W}_{\!S} \leftarrow \bm{W}_{\!S} \odot (\bm{M} - \bm{E}), \  \bm{W}_{\!S} \leftarrow \bm{W}_{\!S} \odot (\bm{M} + \bm{G}).
  \end{gathered}
\vspace{-2pt}
\end{equation}
By doing so, the sparsity of each spatial group is kept steady. Moreover, the weights can be adjusted dynamically, allowing for improved local characteristics. Please refer to Fig.\!~\ref{fig:SDS} for details.$_{\!}$ Once training is complete, mask $\bm{M}$ in spatial groups is also recorded. Unlike previous methods~\cite{liu2022more}, we employ the \textit{Spatial-wise Group Convolution} and partition dynamic sparse processes into separate spatial groups. Pruning and growth are conducted independently within each group without interfering with each other.

We can assume that weights are changing factors over time. Then, removing the least important weights is akin to the selection phase in natural evolution. Alternatively, the random addition of new weights is analogous to the alteration stage of evolutionary selection. This phenomenon is also analogous to a biological process in the brain during sleep, known as synaptic shrinking. Researchers found that the weakest neural link in the brain weakens during slumber, while the vital neural connections remain how they are before. This shows that one of the main roles of sleeping is to reset the overall synaptic strength~\cite{diering2017homer1a,de2017ultrastructural}.

\setlength{\dblfloatsep}{6pt}%
\begin{figure*}[t]
  \centering
  \vspace{-6pt}
      \includegraphics[width=0.99 \linewidth]{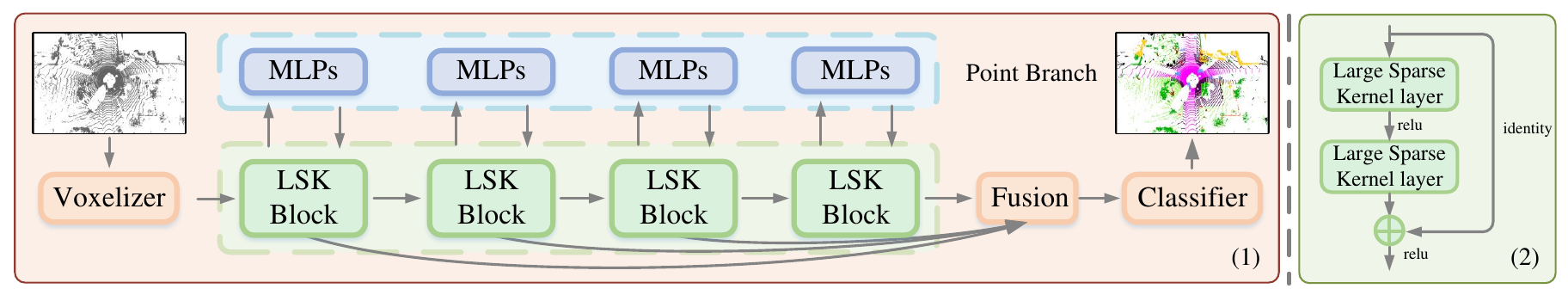}
\vspace{-12pt}
\captionsetup{font=small}
\caption{\small \textbf{LSK3DNet and LSK Block.}~(1)$_{\!}$ {LSK3DNet}: Point clouds are fed into voxelizer for voxel-wise features. Then we extract features with LSK Block and Point Branch. The final prediction is a point-wise output. Here MLPs is the standard Multi-Layer Perceptrons.$_{\!}$ (2)$_{\!}$ {Large Sparse Kernel Block} ({LSK Block}) (Sec.\!~\ref{sec:Objective}).}
\label{fig:framework}
\vspace{-16pt}
\end{figure*}

\subsection{Channel-wise Weight Selection ({CWS})}
\vspace{-1pt}
\label{sec:Reconstitution}
Despite SLaK~\cite{liu2022more} employs DST to enlarge the 2D kernel size, it mitigates the performance degradation caused by sparsity in the way of expanding model width.$_{\!}$ It strives to strike a balance between sparsity and width.$_{\!}$ However, expanding width leads to the issue of increased computational burden.$_{\!}$ Therefore, SLaK~\cite{liu2022more} faces the dilemma that larger sparsity and smaller model size cannot be achieved simultaneously. Instead of naively increasing network width, we propose {CWS} to decouple sparsity and width. It selectively chooses the most salient channels and obtains improved performance while keeping the original model size during inference.

CWS operates in an online mode with a sorting frequency of $f_s$. A single weight sorting cycle involves multiple iterations of \textit{Sparse Weight Pruning and Growth}, \ie, each cycle of weight sorting is a multiple of the adaptation rate $f_a$.$_{\!}$ The width factor $w$ determines the extent to which the number of channels is augmented, \ie, extending $D$-dimension to $w\!\times\!D$-dimension.$_{\!}$ Once the expanded model has been created, we consider the $D$-dimensional small model to be embedded within the $w\!\times\!D$-dimensional augmented model~\cite{cai2022network}. When the number of training iterations reaches the sorting condition, the channels are sorted based on their L1 value (\ie, {\color{apurple}SortChannels}), prompting the model to focus on the most relevant channels. During validation, we select the most important $D$-dimensional channels from $w\!\times\!D$-dimensional channels (\ie, {\color{apurple}SelectChannels}). The pseudo-code for the entire algorithm is provided in Alg.\!~\ref{alg:algorithm1}. This operation effectively achieves higher performance while maintaining the parameters within the expected size. This is especially important in deployments where memory usage and computational efficiency are critical factors.

\begin{table*}[t]
\captionsetup{font=small}
\caption{\small Results of LSK3DNet on SemanticKITTI~\cite{behley2019semantickitti} single-scan \textit{test} (Sec.\!~\ref{sec:segmentation}). Regarding input data format, P denotes points, V represents voxelizations, R signifies range images, and 2DPASS incorporates additional 2D data. ``ms/s" means ``milliseconds per scene".}
\label{tab:kitti_seg}
\centering
\small
\setlength\tabcolsep{4pt}
\renewcommand\arraystretch{1.1}
\resizebox{\linewidth}{!}{
\begin{tabular}{r|c||c||c|c|c|c|c|c|c|c|c|c|c|c|c|c|c|c|c|c|c||c}
\hline
\textbf{Methods}&\rotatebox{90}{Input} & \rotatebox{90}{\textbf{mIoU}} & \rotatebox{90}{road}& \rotatebox{90}{sidewalk}& \rotatebox{90}{parking}& \rotatebox{90}{other-ground~}& \rotatebox{90}{building}& \rotatebox{90}{car}& \rotatebox{90}{truck}& \rotatebox{90}{bicycle}& \rotatebox{90}{motorcycle}& \rotatebox{90}{other-vehicle}& \rotatebox{90}{vegetation}& \rotatebox{90}{trunk}& \rotatebox{90}{terrain}& \rotatebox{90}{person}& \rotatebox{90}{bicyclist}& \rotatebox{90}{motorcyclist}& \rotatebox{90}{fence}& \rotatebox{90}{pole}& \rotatebox{90}{traffic sign} & \rotatebox{90}{speed (ms/s)} \\
\hline
\hline
PointNet++~\cite{qi2017pointnet++}&P&20.1& 72.0 &41.8 &18.7& 5.6 &62.3 &53.7 &0.9 &1.9 &0.2& 0.2& 46.5 &13.8 &30.0 &0.9 &1.0 &0.0 &16.9 &6.0 &8.9 &5900\\
TangentConv~\cite{tatarchenko2018tangent}&R &40.9 &83.9 &63.9 &{33.4} &{15.4} &{83.4} &{90.8} &15.2&{2.7}& 16.5 &12.1 &79.5 &49.3 &58.1 &23.0 &28.4 &{8.1} &{49.0} &35.8 &28.5 &3000\\
PolarNet~\cite{zhang2020polarnet}&R&54.3 &90.8 &74.4 &61.7 &21.7 &90.0 &93.8 &22.9 &40.3 &30.1 &28.5 &84.0 &65.5 &67.8 &43.2 &40.2 &5.6 &61.3 &51.8 &57.5 & 62\\
RandLA-Net~\cite{hu2019randla}&P &55.9 &90.5 &74.0 &61.8 &24.5 &89.7 &94.2 &43.9 &29.8 &32.2 &{39.1} &83.8 &63.6 &68.6 &48.4 &47.4 &9.4 &60.4 &51.0 &50.7 & 880\\
SqueezeSegV3~\cite{xu2020squeezesegv3}&R &55.9 &91.7 &74.8 &63.4 &26.4 &89.0 &92.5 &29.6 &38.7 &36.5 &33.0 &82.0 &58.7 &65.4 &45.6 &46.2 &20.1 &59.4 &49.6 &58.9  & 238\\
KPConv~\cite{thomas2019kpconv}&P &58.8 &{90.3} &72.7 &{61.3} &31.5 &90.5 &95.0 &33.4 &30.2 &42.5 &44.3 &84.8 &69.2 &69.1 &61.5 &61.6 &11.8 &64.2 &56.4 &47.4 & - \\
JS3C-Net~\cite{yan2021sparse}&V & 66.0 & 88.9& 72.1& 61.9& 31.9& 92.5& 95.8& 54.3 & 59.3& 52.9& 46.0& 84.5 & 69.8 & 67.9& 69.5 & 65.4 & 39.9 & 70.8 & 60.7 & 68.7 & 471\\
SPVNAS~\cite{tang2020searching}&PV  & 67.0  & 90.2  & 75.4  & 67.6  & 21.8  & 91.6  & 97.2  & 56.6  & 50.6  & 50.4  & 58.0  & 86.1  & 73.4  & 71.0  & 67.4  & 67.1  & 50.3  & 66.9  & 64.3  & 67.3 &259 \\
Cylinder3D~\cite{zhu2021cylindrical}&V  & 68.9  & 92.2  & 77.0  & 65.0  & 32.3  & 90.7  & 97.1  & 50.8  & 67.6  & 63.8  & 58.5  & 85.6  & 72.5  & 69.8  & 73.7  & 69.2  & 48.0  & 66.5  & 62.4  & 66.2 & 131 \\
RPVNet~\cite{xu2021rpvnet}&RPV  & 70.3  & 93.4  & 80.7  & 70.3  & 33.3  & 93.5  & 97.6  & 44.2  & 68.4  & 68.7  & 61.1  & 86.5  & 75.1 & 71.7  & 75.9  & 74.4  & 43.4  & 72.1  & 64.8  & 61.4  & 168 \\
(AF)$^2$-S3Net~\cite{cheng20212}&V  & 70.8  & 92.0  & 76.2  & 66.8  & 45.8  & 92.5  & 94.3  & 40.2 & 63.0 & 81.4 & 40.0  & 78.6  & 68.0  & 63.1  & 76.4  & 81.7  & 77.7  & 69.6  & 64.0  & 73.3 & - \\
PV-KD~\cite{hou2022point}&V &71.2& 91.8 & 77.5 & 70.9 & 41.0 & 92.4 & 97.0 & 53.5 & 67.9 & 69.3 & 60.2 & 86.5 & 73.8 & 71.9 & 75.1 & 73.5 & 50.5 & 69.4 & 64.9 & 65.8& 76 \\
2DPASS~\cite{yan20222dpass}&PV  & 72.9 & 89.7  & 74.7  & 67.4  & 40.0  & 93.5  & 97.0  & 61.1  & 63.6  & 63.4  & 61.5  & 86.2  & 73.9  & 71.0  & 77.9  & 81.3  & 74.1  & 72.9  & 65.0 &70.4 & 62 \\
SphereFormer~\cite{lai2023spherical} & V & 74.8 &91.8& 78.2& 69.7& 41.3& 93.8& 97.5& 59.6& 70.1& 70.5& 67.7& 86.7& 75.1& 72.4& 79.0& 80.4& 75.3& 72.8& 66.8& 72.9& 123 \\
\hline
\hline
LSK3DNet&PV & \bf{75.6} & 92.2 & 78.9& 70.2&41.8&92.7&97.3&61.0&71.4&75.6&64.2&86.4&72.7&71.9&81.2&80.6&85.2&72.0&67.0&74.6& 89\\
\hline
\end{tabular}}
\vspace{-12pt}
\end{table*}

\subsection{Network Architecture}
\vspace{-1pt}
\label{sec:Objective}
\noindent\textbf{Segmentation Baseline.}~U-Net type 3D networks (such as SparseConvNet~\cite{graham20183d} and MinkUNet~\cite{choy20194d}) employ aggressive downsampling (\ie, regular sparse convolution) to increase the receptive field but at the cost of reduced resolution. However, with a low resolution, several points or tiny objects could be combined into one single grid and become indistinguishable~\cite{tang2020searching}. So SPVCNN~\cite{tang2020searching} is equipped with a high-resolution point-based branch. Subsequently, 2DPASS~\cite{yan20222dpass} further modifies multi-representation branches by omitting the regular sparse convolution. This is because it dilates all sparse features and blurs valuable information, increasing the burden for following layers~\cite{chen2022focal}. Only submanifold sparse convolution is employed for feature extraction in Modified SPVCNN~\cite{yan20222dpass}. By limiting the output feature positions to the input positions, submanifold sparse convolution is able to circumvent the computation burden. We use Modified SPVCNN as a baseline for our segmentation network.$_{\!}$ This 3D network is compact yet powerful and can generate high-resolution representations from sparse point clouds in large-scale scenes.$_{\!}$ However, one challenge we face is that the restricted area of submanifold sparse convolution limits the information flow and makes it difficult to distinguish different spatial characteristics of the scene. To address this challenge, we introduce LSK Block, which stands for Large Sparse Kernel Block. This block increases the kernel sizes of submanifold sparse convolution and expands the receptive field to facilitate information flow. LSK Block has a standard residual structure that adds the output of identity mapping to that of two stacked large kernel convolutions (Fig.\!~\ref{fig:framework}\!~(2)). Our network does not need the parallel convolutional branch with dilatation convolution~\cite{chen2022scaling}. The details of our segmentation network architecture are listed below:

\setlength{\textfloatsep}{4pt}%
\begin{algorithm}[tb]
\captionsetup{font=small}
\caption{\small Pseudo-code of SDS and CWS}
\label{alg:algorithm1}
\begin{algorithmic}[1]
\small
\STATE set adaptation frequency $f_a$, sparsity $s$;
\STATE set sorting frequency $f_s$ and width factor $w$;
\STATE expand model width form $D$ dimension to $w\!\times\!D$ dimension;
\STATE initialize dense model $\bm{W}_{\!D}^{w\!\times\!D}$;
\STATE initialize sparse layers $\bm{W}_{\!S}^{w\!\times\!D}$;
\FOR{each training iteration $i$}
    \STATE $\bm{W}_{\!S}^{w\!\times\!D}\leftarrow${\color{apurple}NormalTraining}$(\bm{W}_{\!S}^{w\!\times\!D})$; \\
    \IF{($i\%{f_a}$) equals to 0}
        \STATE $\bm{W}_{\!S}^{w\!\times\!D}\leftarrow$ $\bm{W}_{\!S}^{w\!\times\!D} \odot (\bm{M} - \bm{E})$; \\
        \STATE $\bm{W}_{\!S}^{w\!\times\!D}\leftarrow$ $\bm{W}_{\!S}^{w\!\times\!D} \odot (\bm{M} + \bm{G})$; \\
    \ENDIF
    \IF{($i\%{f_s}$) equals to 0}
        \STATE $\bm{W}_{\!S}^{w\!\times\!D}\leftarrow${\color{apurple}SortChannels}$(\bm{W}_{\!S}^{w\!\times\!D})$;
    \ENDIF    
    \IF{validation}
        \STATE $\bm{W}_{\!S}^{w\!\times\!D}\leftarrow${\color{apurple}SortChannels}$(\bm{W}_{\!S}^{w\!\times\!D})$;
        \STATE $\bm{W}_{\!S}^{D}\leftarrow${\color{apurple}SelectChannels}$(\bm{W}_{\!S}^{w\!\times\!D})$;
        \STATE {\color{apurple}NormalValidation}$(\bm{W}_{\!S}^{D})$;
    \ENDIF
\ENDFOR
\end{algorithmic}
\end{algorithm}

\begin{itemize}[leftmargin=*]
   \setlength{\itemsep}{0pt}
   \setlength{\parsep}{-2pt}
   \setlength{\parskip}{-0pt}
   \setlength{\leftmargin}{-10pt}
   \item \noindent\textbf{Segmentation Network on SemanticKITTI~\cite{behley2019semantickitti}.}~LSK- 3DNet (See Fig.\!~\ref{fig:framework}\!~(1)) divides the entire scene into voxels, each with a size of 5 cm.$_{\!}$ It has four scales of \{2, 4, 8, 16\}.$_{\!}$ The hiden size $D$ of the entire network is 64. We deploy LSK Blocks in \textit{SparseBasicBlock}\footnote{\url{https://github.com/yanx27/2DPASS}}. Following~\cite{hu2020randla,zhu2021cylindrical,yan20222dpass}, we employ the weighted cross-entropy loss to optimize point accuracy and utilize the lovasz-softmax~\cite{berman2018lovasz} loss to maximize the intersection-over-union.

   \item \noindent\textbf{Segmentation Network on ScanNet v2~\cite{dai2017scannet}.}~This segmentation network has the same architecture as that on SemanticKITTI~\cite{behley2019semantickitti}. The entire scene is split with a voxel size of 2 cm, with scales of \{2, 4, 8, 16, 16\}. The hidden size $D$ is 128. Following~\cite{wu2022point}, we take the weighted cross-entropy loss as the objective function.
\end{itemize}

\noindent\textbf{Detection Network on KITTI.}~We take Voxel R-CNN~\cite{deng2021voxel} as a baseline. Specifically, we retain the backbone of Voxel R-CNN and substitute plain sparse convolutional block with LSK Block in the first three stages. Other settings remain the same as \cite{chen2022scaling}.$_{\!}$ The input spatial shape is \{1600, 1408, 41\}, and the channels of stages are \{16, 16, 32, 64, 64\}.


\vspace{-3pt}
\section{Experiment}
\vspace{-1pt}

\subsection{Setups and Implementations}
\vspace{-1pt}

\noindent\textbf{Dataset.}~To evaluate our method, we perform experiments 

\noindent on three benchmark datasets and five tracks:
\begin{itemize}[leftmargin=*]
   \setlength{\itemsep}{0pt}
   \setlength{\parsep}{-2pt}
   \setlength{\parskip}{-0pt}
   \setlength{\leftmargin}{-10pt}
   \item \textbf{SemanticKITTI}~\cite{behley2019semantickitti} has $43,\!551$ traffic point cloud scenes with fine annotations, split into $19,\!130/4,\!071/20,\!350$ scenes for \texttt{train}/\texttt{val}/\texttt{test}. The dataset has $28$ semantic classes, but only $19$ classes are used for single-scan track and $25$ classes for multi-scan track.
   
   \item \textbf{ScanNet v2}~\cite{dai2017scannet} contains $1,\!201/312/100$ indoor scenes for \texttt{train}, \texttt{val}, and \texttt{test} splits, respectively. There are $20$ semantic categories for ScanNet20 track and $200$ categories for ScanNet200 track.
      
   \item \textbf{KITTI}~\cite{geiger2013vision} has $7,\!481/7,\!518$ samples for \texttt{train} and \texttt{test}.$_{\!}$ We follow the frequently used \texttt{train}/\texttt{val} split~\cite{meng2020weakly} to divide the training samples into \textit{train} split ($3,\!712$ samples) and \textit{val} split ($3,\!769$ samples).
\end{itemize}

\begin{table}[t]
\captionsetup{font=small}
\caption{\small Results of LSK3DNet on SemanticKITTI~\cite{behley2019semantickitti} multi-scan \textit{test} set. The arrow below classes indicate moving classes (Sec.\!~\ref{sec:segmentation}).}
\label{tab:kitti-multi}
\centering
\small
\setlength\tabcolsep{4pt}
\renewcommand\arraystretch{1.0}
\resizebox{\linewidth}{!}{
\begin{tabular}{r|c||c|c||c|c|c|c|c|c}
\hline 
Method&\rotatebox{90}{Input}  &\rotatebox{90}{\bf{mIoU}}&\rotatebox{90}{\bf{Acc}}&\rotatebox{90}{$\underrightarrow{\text{car}}$}&
\rotatebox{90}{$\underrightarrow{\text{truck}}$}&\rotatebox{90}{$\underrightarrow{\text{other-vehicle }}$}&\rotatebox{90}{$\underrightarrow{\text{person}}$}&\rotatebox{90}{$\underrightarrow{\text{bicyclist}}$}&\rotatebox{90}{$\underrightarrow{\text{motorcyclist}}$} \\
\hline 
\hline
LatticeNet~\cite{rosu2019latticenet}&P & 45.2  & 89.3 & 54.8   & 3.5   & 0.6 & 49.9  & 44.6 & 64.3  \\
TLSeg~\cite{duerr2020lidar}&R & 47.0  & 89.6 & 68.2& 2.1    & 12.4 & 40.4   & 42.8  & 12.9  \\
KPConv~\cite{thomas2019kpconv}&P & 51.2  & 89.3  & 69.4& 5.8   & 4.7 & 67.5 & 67.4  & 47.2  \\
Cylinder3D~\cite{zhu2021cylindrical}&V & 52.5  & 91.0  & 74.9  & 0.0  & 0.1 & 65.7 & 68.3   & 11.9  \\
(AF)$^2$-S3Net~\cite{cheng20212}&V & 56.9  & 88.1 & 65.3  & 5.6 & 3.9 & 67.6 & 66.4 & 59.6  \\
2DPASS~\cite{yan20222dpass}&PV &  62.4  &  91.4  &82.1 & 16.1 & 3.8 & 80.3 & 71.2  & 73.1  \\
\hline
\hline
LSK3DNet&PV &  \bf{63.4} & \bf{92.2} & 84.4 & 7.2 & 40.9 & 77.4 & 69.9 & 72.1\\
\hline
\end{tabular}}
\end{table}

\noindent\textbf{Training and Testing Details.}~We utilize AdamW optimizer with OneCycleLR scheduler starting with a learning rate of 5e-3. Data augmentation is also employed, such as random flipping, scaling, rotation around the gravity axis, spatial translation. We apply instance CutMix~\cite{xu2021rpvnet} and Test Time Augmentation (TTA)~\cite{yan20222dpass} to the SemanticKITTI \textit{test} benchmark, and enhance the model with extra training epochs.

\noindent\textbf{Metrics.}~We employ mean class intersection over union (mIoU) and overall accuracy (Acc) metrics as the evaluation criterion for 3D semantic segmentation tasks, as outlined in~\cite{behley2019semantickitti}. What is more, we calculate Average Precision (AP) by recalling 11 positions for 3D object detection.

\begin{table}[t]
\captionsetup{font=small}
\caption{\small Results of LSK3DNet and other tate-of-the-art methods on ScanNet v2~\cite{dai2017scannet} (Sec.\!~\ref{sec:segmentation}). S-Net stands for ScanNet.}
\label{tab:scannetv2}
\centering
\small
\setlength\tabcolsep{8pt}
\renewcommand\arraystretch{1.1}
\resizebox{\linewidth}{!}{
\begin{tabular}{r|c||c|c|c}
\hline
\multirow{2}{*}{Method} & \multirow{2}{*}{\rotatebox{90}{Input  }} & 
 S-Net20&S-Net20& S-Net200 \\
&&\textit{Val} & \textit{Test} & \textit{Val}\\
\hline
\hline
PointNet++~\cite{qi2017pointnet++}             & P & 53.5          & 55.7  &  -   \\
3DMV~\cite{dai20183dmv}                        & P & -             & 48.4  &  -   \\
PanopticFusion~\cite{narita2019panopticfusion} & P & -             & 52.9  &  -   \\
PointCNN~\cite{li2018pointcnn}                 & P & -             & 45.8  &  -   \\
PointConv~\cite{wu2019pointconv}               & P & 61.0          & 66.6  &  -   \\
JointPointBased~\cite{chiang2019unified}       & P & 69.2          & 63.4  &  -   \\
PointASNL~\cite{yan2020pointasnl}              & P & 63.5          & 66.6  &  -   \\
SegGCN~\cite{lei2020seggcn}                    & P & -             & 58.9  &  -   \\
RandLA-Net~\cite{hu2020randla}                 & P & -             & 64.5  &  -   \\
KPConv~\cite{thomas2019kpconv}                 & P & 69.2          & 68.6  &  -   \\
JSENet~\cite{hu2020jsenet}                     & P & -             & 69.9  &  -   \\
FusionNet~\cite{zhang2020deep}                 & P & -             & 68.8  &  -   \\
Point Transformer~\cite{zhao2021point}                     & P & 70.6     & -   & -    \\
Fast Point Transformer~\cite{park2022fast} & P & 72.1 & - &-\\
Stratified Transformer~\cite{lai2022stratified} & P & 74.3 & 73.7 &-\\ 
Point Transformer v2~\cite{wu2022point}  & P & 75.4 & 75.2 & 31.9\\
\hline
\hline
SparseConvNet~\cite{graham20183d,wu2022point}  & V & 69.3          & 72.5  &  28.8   \\
MinkUNet~\cite{choy20194d}                     & V & 72.2          & 73.6  &  -   \\
LargeKernel3D~\cite{chen2022scaling}           & V & 73.5          & 73.9     &   -  \\
LSK3DNet                            & PV & \bf{75.7} & \bf{75.5}& \bf{33.1}\\
\hline
\end{tabular}}
\end{table}

\begin{table}[h] 
\captionsetup{font=small}
\caption{\small Results of 3D object detection methods on the car class of KITTI \textit{val} set~\cite{geiger2013vision} (Sec.\!~\ref{sec:detection}).}
\label{tab:kittival}
\centering
\small
\setlength\tabcolsep{10.4pt}
\renewcommand\arraystretch{1.0}
\resizebox{\linewidth}{!}{
\begin{tabular}{r|c||c|c|c}
\hline
\multirow{2}{*}{Method} & \multirow{2}{*}{\rotatebox{90}{Input  }}&        
\multicolumn{3}{c}{3D AP~(IoU=0.7)}\\
\cline{3-5}
&& Easy & Moderate & Hard \\
\hline   
\hline
VoxelNet~\cite{qi2019deep}               & V               & 81.97         & 65.46         & 62.85         \\
PointPillars~\cite{lang2019pointpillars}   &R      & 86.62         & 76.06         & 68.91         \\
SECOND~\cite{yan2018second}                 & V                       & 88.61         & 78.62         & 77.22         \\
Point R-CNN~\cite{shi2019pointrcnn}         & P        & 88.88         & 78.63         & 77.38         \\
Part-$A^2$~\cite{shi2020points}        &V       & 89.47         & 79.47         & 78.54         \\
PV-RCNN~\cite{shi2020pv}              &PV    & 89.35         & 83.69         & 78.70         \\
Focals Conv~\cite{chen2022focal}       & V             & 89.52         & 84.93         & 79.18         \\ 
Voxel R-CNN~\cite{deng2021voxel}        & V          & 89.41         & 84.52         & 78.93         \\
\hline 
\hline
LargeKernel3D~\cite{chen2022scaling} &V &89.52 &85.07 &79.32\\ 
LSK3DNet & V& \bf{90.16}&\bf{85.61}&\bf{79.53}\\
\hline 
\end{tabular}}
\vspace{0.5pt}
\end{table}

\subsection{3D Semantic Segmentation}
\vspace{-1pt}
\label{sec:segmentation}
\noindent\textbf{Results on SemanticKITTI.}~We test LSK3DNet on both single-scan and multi-scan tracks~\cite{behley2019semantickitti}.$_{\!}$ Tab.\!~\ref{tab:kitti_seg} presents the quantitative results of SemanticKITTI single-scan track. LSK3DNet outperforms 2DPASS~\cite{yan20222dpass} in terms of both mIoU and IoU scores for most categories.$_{\!}$ Moreover, SDS reduces the model size of naive 3D kernels and CWS keeps the model size within the desired size. So, LSK3DNet has a faster running speed than most prior methods, but it's performance significantly benefits from the large kernels.$_{\!}$ Tab.\!~\ref{tab:kitti-multi} shows the results of SemanticKITTI multi-scan track. The mIoU and Acc are computed over all 25 classes.$_{\!}$ Due to page limitations, we only report the per-class IoUs for moving categories. Under this challenging setting, LSK3DNet surpasses 2DPASS~\cite{yan20222dpass} in both mIoU and Acc.$_{\!}$ Notably, LSK3DNet achieves state-of-the-art performance on both single-scan and multi-scan tracks in SemanticKITTI.$_{\!}$ Surpassing over the SOTA 2D-3D method 2DPASS is more valuable, which utilizes both 2D and 3D data while our LSK3DNet takes as input only the 3D data.$_{\!}$ Visualization results are in the supplementary materials.

\noindent\textbf{Results on ScanNet V2.}~We compare LSK3DNet with previous state-of-the-art methods on ScanNet v2~\cite{dai2017scannet}, a large-scale dataset for 3D indoor scene segmentation. ScanNet v2 has two tracks: ScanNet20 and ScanNet200, which use 20 and 200 semantic classes respectively. Tab.\!~\ref{tab:scannetv2} presents the quantitative results of our model and other methods on both tracks. LSK3DNet achieves higher performance than previous methods in both tracks.$_{\!}$ Our performance is superior to transformer-based methods~\cite{zhao2021point,park2022fast,lai2022stratified,wu2022point}, including Point Transformer v2. Moreover, LSK3DNet has a clear advantage over LargeKernel3D~\cite{chen2022scaling}, improving the mIoU by 2.2\% and 1.6\% on the ScanNet v2 \textit{val} and \textit{test} set, respectively.

\subsection{3D Object Detection}
\vspace{-1pt}
\label{sec:detection}
We have also evaluated the detection performance of LSK- 3DNet on the \textit{val} split for car category, as shown in Tab.\!~\ref{tab:kittival}.$_{\!}$ We report the average precision metric for a 3D bounding box with 11 recall positions~(Recall 11). Based on Voxel R-CNN~\cite{deng2021voxel}, we showcase the effectiveness of the LSK Block by comparing it with LargeKernel3D~\cite{chen2022scaling}, a previous 3D large-kernel method. LSK3DNet achieves better results in three difficulty levels, compared with~\cite{deng2021voxel,chen2022scaling}. Visualization comparison can be found in the supplementary material.

\subsection{Ablation Studies}
\vspace{-1pt}
\label{sec:ablation}
We first explore the effect of kernel size on performance, and then the overall architecture design. Next, we explain the hyperparameter choices of SDS and CWS.$_{\!}$ All ablation experiments are performed on the \textit{val} set of SemancKITTI.

\noindent\textbf{Kernel Size.}~We use Modified SPVCNN as the baseline and then explore 3D kernel sizes under two settings: naive dense large kernel and our LSK3DNet. In the former setting, the dense 3D kernel is straightforwardly expanded. In the latter one, LSK3DNet is trained simultaneously with SDS and CWS, meaning that our LSK3DNet can learn a large sparse kernel model from the beginning.

\begin{figure}[t]
  \centering
    \includegraphics[width=1.01 \linewidth]{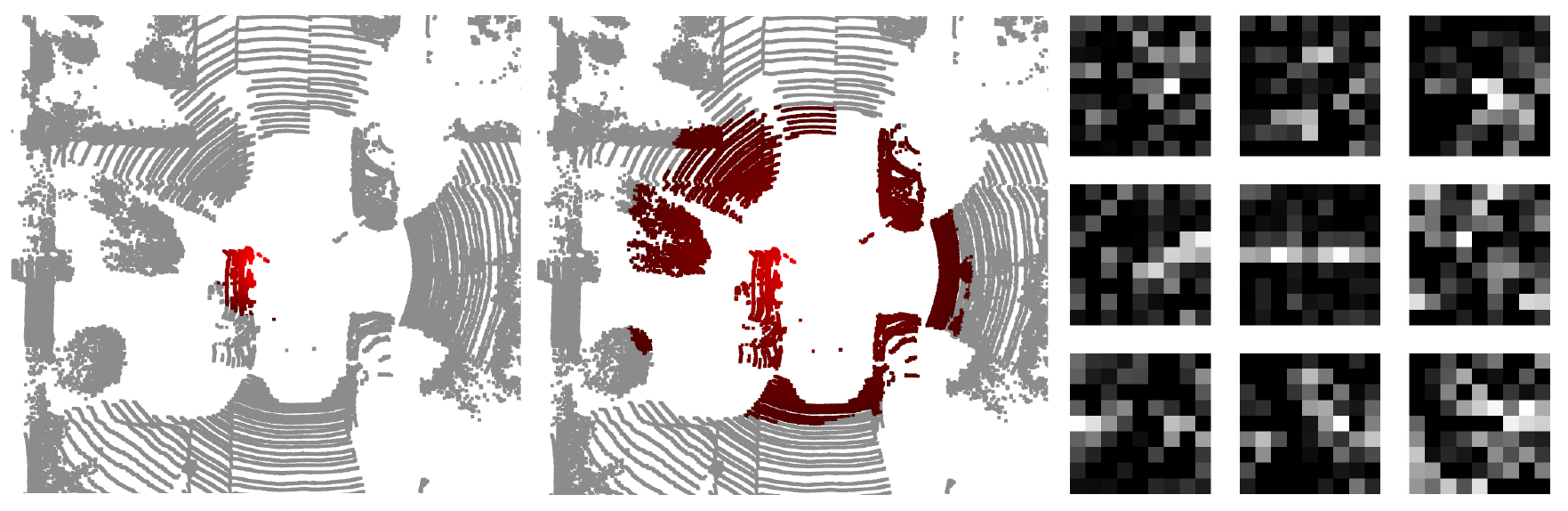}
    \put(-211,-7){\scalebox{.75}{Baseline}}
    \put(-136,-7){\scalebox{.75}{LSK3DNet}}
    \put(-69,-7){\scalebox{.75}{Kernel visualization}}
\vspace{-7pt}
\captionsetup{font=small}
\caption{\small Effective Receptive Fields (ERFs) of Baseline and LSK- 3DNet. LSK3DNet has a larger ERF size. Additionally, we provide visualization of learned sparse kernels, where all weight values have taken the absolute value and normalization operations. The black areas indicate positions with weight values of zero.}
\label{fig:erf}
\end{figure}

The baseline does not use aggressive downsampling which is common in most U-type 3D networks~\cite{graham20183d,choy20194d}. However, submanifold sparse convolution with small kernels may lose important information flow, especially for the spatially disconnected features. We solve this problem by enlarging the size of the 3D convolution kernel to obtain a large receptive field. This agrees with the performance improvement in Tab.\!~\ref{tab:Ablation-KernelSize}, when the kernel size increases. In contrast to this observation, LargeKernel3D ~\cite{chen2022scaling} shows the opposite trends; we empirically attribute this to the fact that LargeKernel3D is based on U-type 3D networks (Sec.\!~\ref{sec:Objective}). For a better understanding of Effective Receptive Fields (ERFs) size, Fig.\!~\ref{fig:erf} illustrates the comparison between the Baseline and LSK3DNet. We follow the definition of ERFs~\cite{luo2016understanding}, and measure the gradient of input point data to the central point of the features generated in the last layer. Compared to the baseline, the high-contribution points of LSK3DNet are distributed over a larger input range, indicating a larger ERF. Additionally, to enhance comprehension of the learned kernels, we offer kernel visualization in Fig.\!~\ref{fig:erf}, providing insights into sparse training.

Compared to the ``naive dense large kernel" of each kernel size, LSK3DNet exhibits superior performance in its corresponding size. LSK3DNet not only leverages SDS to expand its receptive field but also starts from scratch to learn 3D sparse kernels. Concurrently, CWS widens the network during training while pruning redundant channels to accelerate inference. Furthermore, SDS and CWS address the issues of overparameterization and overfitting that arise from simply increasing the kernel's size and expanding the model's width. The best performance of LSK3DNet is achieved at the $9\!\times\!9\!\times\!9$ size, and the following experiments are conducted with $9\!\times\!9\!\times\!9$ size. Moreover, due to the extra dimension in 3D kernels compared to 2D kernels, naively enlarging the 3D kernel size will lead to a cubically-increasing overhead. Our LSK3DNet can effectively reduce 19.1M parameters and approximately 60\% of computing operations when compared to the naive large kernel network. This aspect significantly enhances the value of SDS and CWS for 3D kernels.

\begin{table}[t]
\captionsetup{font=small}
\caption{\small Segmentation performance of Modified SPVCNN with various large kernel settings. ``Dense'' refers to directly enlarging the kernel size. ``Ours'' refers to training with SDS and CWS (Sec.\!~\ref{sec:ablation}). The unit for ``Param'' is in million (M), ``FLOPs'' is in billion (G), and ``Speed'' is measured in milliseconds per scene.}
\label{tab:Ablation-KernelSize}
\centering
\small
\setlength\tabcolsep{8pt}
\renewcommand\arraystretch{1.1}
\resizebox{\linewidth}{!}{
\begin{tabular}{r|c|c|c|c|c}
\hline 
 & Kernel Size & mIoU &  Param & FLOPs & Speed $\downarrow$\\
\hline 
\hline
\multirow{5}{*}{\shortstack{Dense\\ ($D=64$)}} 
& \(3,3,3\) & 65.2& 1.9 & 78.4 &36 \\
& \(5,5,5\) & 65.6 & 8.4 & 127.8  &57 \\
&\(7,7,7\) & 66.2 & 22.6& 381.2& 65\\
&\(9,9,9\) & 67.5 & 47.9& 1916.3 &93 \\
&\(11,11,11\) & 66.6 &87.4 & 3192.0& 134\\
\hline
\multirow{4}{*}{\shortstack{Dense \\ ($1.8\!\times\!D$)}} 
& \(5,5,5\) & 66.2 & 27.0 & 444.0 & 84 \\
& \(7,7,7\) & 66.5 & 73.1& 1551.2 & 123\\
&\(9,9,9\) & 67.2 & 154.7& 3030.0 & 177\\
&\(11,11,11\) & 66.3 &  282.1 & 7139.2& 250 \\
\hline
\multirow{4}{*}{\shortstack{Ours \\ ($1.8\!\times\!D$)}} 
& \(5,5,5\) & 66.8 & 5.1 & 203.8 & 46\\
& \(7,7,7\) & 68.1 & 13.6& 249.4 & 63\\
&\(\bf{9,9,9}\) & \bf{70.2} & 28.8& 763.6&89\\
&\(11,11,11\) & 70.1 & 52.5 &1847.4 &116 \\
\hline
\end{tabular}
}
\vspace{-8pt}
\end{table}

\begin{table}[t]
\captionsetup{font=small}
\caption{\small Overall architecture design with $9\!\times\!9\!\times\!9$ size. Here, we report Param, FLOPs, and Speed for the inference model (Sec.\!~\ref{sec:ablation}).}
\label{tab:Ablation-overall}
\centering
\small
\setlength\tabcolsep{8.4pt}
\renewcommand\arraystretch{1.0}
\resizebox{\linewidth}{!}{
\begin{tabular}{r|c|c|c|c}
\hline
Method & mIoU & Param & FLOPs & Speed $\downarrow$\\ \hline
Dense~~($1.0\!\times\!D$) & 67.5  & 47.9& 1916.3 &93\\
Dense~~($1.8\!\times\!D$) & 67.2  & 154.7& 3030.0 &177\\
\hline
SDS~~($1.8\!\times\!D$) & 69.3 & 93.0&2201.4 &173\\
CWS~~($1.8\!\times\!D$) & 69.1 & 47.9& 1916.3 &93\\
\hline
\bf{LSK3DNet}~~\bf{(}$\bf{1.8\!\times\!D}$\bf{)} & \bf{70.2}& \bf{28.8} & \bf{763.6}&\bf{89}\\
\hline
\end{tabular}
}
\vspace{1pt}
\end{table}

\noindent\textbf{Overall Architecture Design.}~In Tab.\!~\ref{tab:Ablation-overall}, Dense ($1.8\!\times\!D$) does not yield improvement due to overparameterization and overfitting. Although SDS maintains a model width of $1.8\!\times\!D$ during inference, it achieves improved performance due to sparse training and the large receptive field. CWS effectively addresses overfitting by selecting salient channels, thereby enhancing performance. By combining SDS and CWS, LSK3DNet can achieve the best performance while significantly reducing the model size and computational cost.

\noindent\textbf{Spatial-wise Dynamic Sparsity.}~To control the degree of sparse weight adaptation in LSK3DNet, we have adjusted three hyperparameters: the adaptation frequency $f_a$, sparsity rate $s$, and prune rate $p$ (Sec.\!~\ref{sec:Sparsity}). Adaptation frequency $f_a$ indicates how often we update the sparse weights during training. Sparsity rate $s$ specifies how sparse the 3D large kernels are.$_{\!}$ Prune rate $p$ shows the fraction of updated weights in one adaptation. We have conducted experiments with different values of these hyperparameters and report the results in Tab.\!~\ref{tab:Ablation-speedanalysis} and Tab.\!~\ref{tab:hyper_sparsity}. Based on our empirical findings, we have chosen $f_a\!=\!2000$, $s\!=\!0.4$, and $p\!=\!0.3$ as the optimal settings for LSK3DNet.

\begin{table}[t]
\captionsetup{font=small}
\caption{\small Training and inference speed analysis, speed is measured in milliseconds per scene. The kernels have $9\!\times\!9\!\times\!9$ size, ``T'' and ``I'' represent training and inference speed, respectively (Sec.\!~\ref{sec:ablation}).}
\label{tab:Ablation-speedanalysis}
\centering
\small
\setlength\tabcolsep{4pt}
\renewcommand\arraystretch{1.0}
\resizebox{\linewidth}{!}{
\begin{tabular}{r|c|c|c|c|c|c|c|c|c}
\hline 
Sparsity & \multicolumn{3}{c|}{Dense~~($1.0\!\times\!D$)} & \multicolumn{3}{c|}{SDS~~($1.8\!\times\!D$)} &  \multicolumn{3}{c}{LSK3DNet~~($1.8\!\times\!D$)}\\
\cline{2-10}
s &mIoU &T $\downarrow$&I $\downarrow$& mIoU &T $\downarrow$&I $\downarrow$&mIoU &T $\downarrow$&I $\downarrow$\\
\hline 
\hline
0 &67.5 &451& 93 &- &- &- &-&- &-\\
0.1&- &-&-& 68.5 & 489& 176& 69.1 &491&92\\
0.2&- &-&-& 68.9& 478&  176 &69.7 &482&91\\
\bf{0.4}&- &-&-& 69.3& 463&173 &\bf{70.2} &467&89\\
0.6&- &-&-& 69.0& 450& 171 & 68.8&454 &87\\
0.8&- &-&-& 68.7& 439 & 171&67.2& 443 &87\\
\hline
\end{tabular}}
\vspace{-6.5pt}
\end{table}

\begin{table}[t]  
  \captionsetup{font=small}
  \caption{\small $_{\!}$Ablation$_{\!}$ studies$_{\!}$ of$_{\!}$ \textit{Spatial-wise Dynamic Sparsity} on adaptation frequency $f_a$ and prune rate $p$ (Sec.\!~\ref{sec:ablation}).}
  \label{tab:hyper_sparsity}
  \hspace{2pt}
  \begin{subtable}[t]{.2\textwidth}
    \centering
    \small
  \setlength\tabcolsep{7.5pt}
\renewcommand\arraystretch{0.9}
  \resizebox{0.95\linewidth}{!}{
    \begin{tabular}{r|c}\hline
     Adaptation $f_a$ & mIoU \\ \hline\hline
      100  & 68.8  \\
      1000  & 69.6 \\
      \bf{2000}  & \bf{70.2} \\
      3000 & 70.0\\
      4000  & 69.8 \\
      \hline
    \end{tabular}}
  \end{subtable}
 \hspace{16pt}
  \begin{subtable}[t]{.2\textwidth}
    \centering
    \small
  \setlength\tabcolsep{11.5pt}
\renewcommand\arraystretch{0.9}
  \resizebox{0.95\linewidth}{!}{
    \begin{tabular}{r|c}\hline
     Pruning $p$ & mIoU \\ \hline\hline
      0.10  & 69.9  \\
      0.20  & 70.1  \\
      \bf{0.30}  & \bf{70.2} \\
      0.50  & 70.0 \\
      0.70 & 69.7 \\
      \hline
    \end{tabular}}
  \end{subtable}
  \vspace{-5.5pt}
\end{table}

\noindent\textbf{Channel-wise Weight Selection.}~Another aspect of our method that we investigated is the CWS. This technique involves two hyperparameters: the sorting frequency $f_s$ and the width factor $w$ (Sec.\!~\ref{sec:Reconstitution}). The former determines how frequently we sort and select channels for large kernels during training.$_{\!}$ It is expressed as a multiple of $f_a$, the adaptation frequency, meaning that we perform channel sorting after a certain number of weight adaptation cycles. When $f_s\!=\!6\!\times\!f_a$, the optimal performance is achieved. The latter controls the times of network width compared to the target model. We find that increasing the width factor up to $1.8 \times$ improves the model’s performance, but beyond that point, there is no significant gain. Therefore, we opt for a width of $1.8 \times$ to avoid more extra compute resources that a wider width would require during training.

\begin{table}[t]  
  \captionsetup{font=small}
  \caption{\small $_{\!}$Ablation$_{\!}$ studies$_{\!}$ of$_{\!}$ \textit{Channel-wise Weight Selection} on sorting frequency $f_s$ and width factor $w$ (Sec.\!~\ref{sec:ablation}).}
  \label{tab:hyper_sparsity2}
  \hspace{7pt}
  \begin{subtable}[t]{.2\textwidth}
    \centering
    \small
  \setlength\tabcolsep{8pt}
\renewcommand\arraystretch{0.9}
  \resizebox{0.88\linewidth}{!}{
    \begin{tabular}{r|c}\hline
     Sorting $f_s$ & mIoU \\ \hline\hline
      $1 \times f_a$  & 68.2  \\
      $2 \times f_a$  & 69.1 \\
      $4 \times f_a$  & 69.8 \\
      $\bf{6 \times f_a}$ & \bf{70.2}\\
      $10 \times f_a$  & 69.9 \\
      \hline
    \end{tabular}}
  \end{subtable}
 \hspace{11pt}
  \begin{subtable}[t]{.2\textwidth}
    \centering
    \small
  \setlength\tabcolsep{11.5pt}
\renewcommand\arraystretch{0.9}
  \resizebox{0.95\linewidth}{!}{
    \begin{tabular}{r|c}\hline
     Width $w$ & mIoU \\ \hline\hline
      $1.1 \times$  & 68.4  \\
      $1.5 \times$  & 69.6  \\
      $\bf{1.8 \times}$  & \bf{70.2} \\
      $2.1 \times$  & 69.9 \\
      $2.5 \times$ & 70.1\\
      \hline
    \end{tabular}}
  \end{subtable}
\vspace{1.5pt}
\end{table}

\vspace{-3pt}
\section{Conclusion}
\vspace{-1pt}
We propose \textit{Spatial-wise Dynamic Sparsity} to scale up 3D kernels beyond $9\!\times\!9\!\times\!9$, which prunes the volumetric weight and reduces the parameter size of large kernel layers. Our LSK3DNet can benefit from a large receptive field without increasing the computational cost compared to a naive 3D large kernel. \textit{Channel-wise Weight Selection} expands the model width during training, and then sorts and selects important channels during validation to get a model of the expected size. In this way, we achieve \textit{``using spatial sparse groups, expanding width without more parameters''}. We evaluate our method on the SemanticKITTI and achieve state-of-the-art performance. Our LSK3DNet also surpasses previous 3D large kernel methods on ScanNet v2 and KITTI.

{
\small
\bibliographystyle{ieeenat_fullname}
\bibliography{egbib}
}

\input{LSK3DNet_suppl}

\end{document}

%% file: preamble.tex
\usepackage[numbers,sort&compress]{natbib}

\usepackage[dvipsnames]{xcolor}

\usepackage{epsfig}
\usepackage{graphicx}
\usepackage{amsmath}
\usepackage{amssymb}
\usepackage{times}
\usepackage{multirow}
\usepackage{pifont}
\usepackage{booktabs}
\usepackage{booktabs,multirow}
\usepackage{graphics}
\usepackage{threeparttable}
\usepackage{color}
\usepackage[normalem]{ulem}
\usepackage{multirow}
\usepackage{float}
\usepackage{amsfonts}
\usepackage{bm}
\usepackage{enumitem}
\usepackage{array}
\definecolor{myy}{RGB}{126,95,0}
\definecolor{mygray}{gray}{.9}
\definecolor{bblue}{RGB}{30,80,120}
\definecolor{mygray1}{gray}{.7}
\usepackage{bm}
\usepackage{mathtools}
\usepackage{subcaption}
\usepackage{siunitx}
\usepackage[nopar]{lipsum}
\usepackage[export]{adjustbox}

\usepackage{url}
\usepackage{nicefrac}       
\usepackage{microtype}      
\usepackage{algorithm}
\usepackage{algorithmic}
\usepackage{wrapfig}
\usepackage{arydshln}
\usepackage{dblfloatfix}

\usepackage{soul}
\soulregister\cite7
\soulregister\ref7
\soulregister\pageref7

\newcolumntype{I}{!{\vrule width 1pt}}

\definecolor{ggray}{RGB}{127,127,127}
\definecolor{mygreen}{RGB}{93,174,86}

\definecolor{apurple}{RGB}{102, 77, 166}
\definecolor{agreen}{RGB}{117, 175, 92}
\definecolor{ablue}{RGB}{66, 115, 147}
\definecolor{ablue}{RGB}{66, 115, 147}
\definecolor{ablack}{RGB}{57, 57, 57}
\definecolor{agray}{RGB}{101, 101, 101}

\makeatletter
\newcommand{\thickhline}{%
   \noalign {\ifnum 0=`}\fi \hrule height 1pt
   \futurelet \reserved@a \@xhline
}
\makeatother

\usepackage{caption}
\usepackage{makecell}

\newcommand\figcaption{\def\@captype{figure}\caption}
\newcommand\tabcaption{\def\@captype{table}\caption}

\usepackage{algorithm}
\usepackage{listings}
\usepackage{etoolbox}
\makeatletter
\AfterEndEnvironment{algorithm}{\let\@algcomment\relax}
\AtEndEnvironment{algorithm}{\kern2pt\hrule\relax\vskip3pt\@algcomment}
\let\@algcomment\relax
\newcommand\algcomment[1]{\def\@algcomment{\footnotesize#1}}
\renewcommand\fs@ruled{\def\@fs@cfont{\bfseries}\let\@fs@capt\floatc@ruled
  \def\@fs@pre{\hrule height.8pt depth0pt \kern2pt}%
  \def\@fs@post{}%
  \def\@fs@mid{\kern2pt\hrule\kern2pt}%
  \let\@fs@iftopcapt\iftrue}
\makeatother


%% file: LSK3DNet_suppl.tex
\clearpage

\renewcommand{\thefigure}{S\arabic{figure}}
\renewcommand{\thetable}{S\arabic{table}}
\renewcommand{\thepage}{S\arabic{page}} 
\renewcommand{\thealgorithm}{S\arabic{algorithm}} 

\setcounter{page}{1}
\setcounter{table}{0}
\setcounter{figure}{0}
\setcounter{algorithm}{0}
\maketitlesupplementary

\appendix
To enhance the clarity of the main paper, we include the following sections in this supplementary material. We first introduce the analysis and comparison with previous methods in Sec.\!~\ref{appendix_1}. Ablation studies on spatial group size, Effective Receptive Fields (ERFs), inference speed and FLOPs, and Convolutional Schemes are elaborated in Sec.\!~\ref{appendix_2}. More qualitative and quantitative results are further presented and analyzed in Sec.\!~\ref{appendix_3}. Finally, limitations and societal impact are discussed in Sec.\!~\ref{appendix_4}.

\section{Comparison and Discussion}
\label{appendix_1}
\noindent\textbf{Comparison with LinK~\cite{lu2023link} in $s$ and $r$ Settings.}~Block Based Aggregation in LinK~\cite{lu2023link} divides the entire input space into several non-overlapping blocks. Then, it utilizes block-wise proxy aggregation to extract the corresponding features of proxy nodes.$_{\!}$ Furthermore, the corresponding features of the center are obtained by aggregating the features of the neighboring blocks closest to the center. In LinK, the variable $s$ represents the scale of the block, while $r$ denotes the scale of the block query. Actually, this paradigm is not novel; for example, a similar paradigm has already been employed in 2DPASS~\cite{yan20222dpass}. It first divides the entire input space into non-overlapping voxels (blocks) at four scales: 2, 4, 8, and 16. The features of these voxels are obtained by averaging the features within each voxel. Then, 3D sparse convolution is applied to extract features from adjacent voxels.$_{\!}$ Here, the kernel size of sparse convolution plays the same role as $r$.$_{\!}$ In addition, we also use $s$ to represent the above four scales.$_{\!}$ In Tab.\!~\ref{tab:receptivefieldsize}, we have fairly compared the receptive field sizes of each module for three methods, based on the default hayaparameters of models on the SemanticKITTI validation dataset~\cite{behley2019semantickitti}.$_{\!}$ Specifically, three modules are ELKBlock\footnote{\url{https://github.com/MCG-NJU/LinK}\label{foot:1}} in Link, SPVBlock\footnote{\url{https://github.com/yanx27/2DPASS}} in 2DPASS, and LSK Block in our LSK3DNet. LinK achieves an equivalent receptive field size of $21\times21\times21$ for the four ELKBlocks in LinK. However, the second, third, and fourth LSK Blocks in LSK3DNet clearly exhibit larger equivalent receptive fields.

\begin{table}[t]
\caption{\small \textbf{Receptive field size}. RF represents the receptive field size, with subscripts denoting the index of the scale.}
\label{tab:receptivefieldsize}
\centering
\setlength\tabcolsep{24pt}
\renewcommand\arraystretch{1.0}
\resizebox{\linewidth}{!}{
\begin{tabular}{r|c|c}
\hline 
Size &  $s_1$/$r_1$ & $RF_1$  \\
\hline 
\hline
LinK~\cite{lu2023link}& 3/7 & $21\times21\times21$\\
2DPASS~\cite{yan20222dpass} & 2/3 & $6\times6\times6$\\
LSK3DNet& 2/9 & $18\times18\times18$\\
\hline
Size &  $s_2$/$r_2$ & $RF_2$  \\
\hline 
LinK~\cite{lu2023link}& 3/7 & $21\times21\times21$\\
2DPASS~\cite{yan20222dpass} & 4/3 & $12\times12\times12$\\
LSK3DNet& 4/9 & $36\times36\times36$\\
\hline
Size &  $s_3$/$r_3$ & $RF_3$  \\
\hline 
LinK~\cite{lu2023link}& 3/7 & $21\times21\times21$\\
2DPASS~\cite{yan20222dpass} & 8/3 & $24\times24\times24$\\
LSK3DNet& 8/9 & $72\times72\times72$\\
\hline
Size &  $s_4$/$r_4$ & $RF_4$  \\
\hline 
LinK~\cite{lu2023link}& 3/7 & $21\times21\times21$\\
2DPASS~\cite{yan20222dpass} & 16/3 & $48\times48\times48$\\
LSK3DNet& 16/9 & $144\times144\times144$\\
\hline
\end{tabular}}
\vspace{-1pt}
\end{table}

The scales mentioned in section 3.5 and the convolution kernel size have similar spatial meanings to the $s$ and $r$ settings in LinK. Please note that we only adopted kernel size (i.e., $r$ in LinK~\cite{lu2023link}) as a metric for large kernel design in the main paper, following LargeKernel3D~\cite{chen2022scaling}. Please do not confuse these two settings.

\noindent\textbf{Performance of LinK.}~Upon comparing all the results reported in Link~\cite{lu2023link} and Model Zoo\textsuperscript{\ref {foot:1}}, we notice that the highest performance of 67.7\% is achieved with a $(2\times3)^3$ receptive field size, surpassing other sizes such as $(3\times7)^3$ and $(3\times5)^3$. Interestingly, the $(2\times3)^3$ receptive field size appears noticeably smaller than $(3\times7)^3$. It turns out that Link~\cite{lu2023link} achieves its stronger performance with a smaller receptive field.

\noindent\textbf{Kernel Size of LargeKernel3D on ScanNet v2~\cite{dai2017scannet}.}~The claim that “the performance of LargeKernel3D~\cite{chen2022scaling} drops when scaling up the kernel size over $7\!\times\!7\!\times\!7$” is concluded from Table S - 9\footnote{\url{https://arxiv.org/pdf/2206.10555v1.pdf}}.

\section{Ablation Studies}
\label{appendix_2}

\noindent\textbf{Spatial Group Size.}~We report the results of the ablation study on the kernel size selection in Tab.\!~\ref{tab:spatialgroupsize}.$_{\!}$ \textit{kernel size} means the absolute kernel size. \textit{group} means the number of groups to split kernels.$_{\!}$ We conduct the experiment on LSK3DNet on the SemanticKITTI dataset, with the same training hyper-parameters in the main paper. For $3\!\times\!3\!\times\!3$ \textit{group} and $9\!\times\!9\!\times\!9$ \textit{kernel size}, every group has \{3,3,3\} divisions in each dimension. Moreover, $9\!\times\!9\!\times\!9$ is split into $5\!\times\!5\!\times\!5$ with \{2,2,1,2,2\} divisions.$_{\!}$ For $13\!\times\!13\!\times\!3$ \textit{kernel size}, $3\!\times\!3\!\times\!3$ \textit{group} has \{4,5,4\} divisions in 13-dimensional axis, and similarly, \{2,3,3,3,2\} divisions for $5\!\times\!5\!\times\!3$ \textit{group}, \{2,2,2,1,2,2,2\} divisions for $7\!\times\!7\!\times\!3$ \textit{group}. We finally chose the $9\!\times\!9\!\times\!9$ \textit{kernel size} and $3\!\times\!3\!\times\!3$ \textit{group} based on our empirical results. We speculate that this is because each group has a $3\!\times\!3\!\times\!3$ size, showing more parameter space to explore than other \textit{group} settings. Previous studies have shown that exploring a large parameter space is important for dynamic sparse training~\cite{jayakumar2020top,raihan2020sparse,liu2021selfish}. In addition, the kernel size of $13\!\times\!13\!\times\!3$ does not achieve better performance, being affected by overparameterization and overfitting issues.

\begin{table}[t]
\caption{\small \textbf{Ablation studies} on spatial group size.}
\label{tab:spatialgroupsize}
\centering
\setlength\tabcolsep{27pt}
\renewcommand\arraystretch{1.0}
\resizebox{\linewidth}{!}{
\begin{tabular}{r|c|c}
\hline 
\textit{kernel size} &  \textit{Group} & mIoU (\%) \\
\hline 
\hline
\multirow{2}{*}{$9\!\times\!9\!\times\!9$} 
& $3\!\times\!3\!\times\!3$ & 70.2\\
&$5\!\times\!5\!\times\!5$ & 69.1 \\
\hline
\multirow{3}{*}{$13\times13\times3$} 
&$3\!\times\!3\!\times\!3$ & 69.9\\
&$5\!\times\!5\!\times\!3$ & 69.5 \\
&$7\!\times\!7\!\times\!3$ & 68.9 \\
\hline
\end{tabular}}
\vspace{-4pt}
\end{table}

\begin{table}[t]
\caption{\small \textbf{Ablation studies} on \textit{Depth-wise Group Convolution} and \textit{Spatial-wise Group Convolution} on ScanNet v2~\cite{dai2017scannet} and SemanticKITTI~\cite{behley2019semantickitti}.}
\label{tab:groupconvolution}
\centering
\setlength\tabcolsep{4.5pt}
\renewcommand\arraystretch{1.0}
\resizebox{\linewidth}{!}{
\begin{tabular}{r|c|c}
\hline
Methods    &   Kernel  & mIoU (\%) \\ \hline\hline
\multicolumn{3}{c}{\textit{ScanNet v2}}\\ \hline
\multirow{4}{*}{\makecell[r]{MinkowskiNet-14~\cite{chen2022scaling}\\+ {\em depth-wise} conv \\+ {\em spatial-wise} conv \\+ {\em spatial-wise} conv}} & $7\!\times\!7\!\times\!7$  & 68.6 \\ & $7\!\times\!7\!\times\!7$ & 56.4 \\
& $7\!\times\!7\!\times\!7\to3\!\times\!3\!\times\!3$ & 69.7\\
& $9\!\times\!9\!\times\!9\to7\!\times\!7\!\times\!7$ & 69.6\\
\hline
\multirow{3}{*}{\makecell[r]{MinkowskiNet-34~\cite{chen2022scaling}\\+ {\em depth-wise} conv \\+ {\em spatial-wise} conv }} & $7\!\times\!7\!\times\!7$  & 68.6 \\ & $7\!\times\!7\!\times\!7$ & 68.7 \\
& $7\!\times\!7\!\times\!7\to3\!\times\!3\!\times\!3$ & 73.2\\
\hline
\multirow{3}{*}{\makecell[r]{Modified SPVCNN\\+ {\em depth-wise} conv (SDS and CWS) \\+ {\em spatial-wise} conv (SDS and CWS)}} & $9\!\times\!9\!\times\!9$  & 72.4 \\
&  $9\!\times\!9\!\times\!9$   & 67.0 \\
& $9\!\times\!9\!\times\!9$ & 75.7\\
\hline
\multicolumn{3}{c}{\textit{SemanticKITTI}}\\ \hline
\multirow{3}{*}{\makecell[r]{Modified SPVCNN\\+ {\em depth-wise} conv (SDS and CWS) \\+ {\em spatial-wise} conv (SDS and CWS) }} & $9\!\times\!9\!\times\!9$  & 67.5 \\
&  $9\!\times\!9\!\times\!9$   & 65.4 \\
& $9\!\times\!9\!\times\!9$ & 70.2\\
\hline
\end{tabular}}
\end{table}

\noindent\textbf{Inference Speed and FLOPs.}~As described in \S{\color{red}3.3} and Sec.\!~\ref{appendix_4}, SDS involves unstructured sparsity, whose GPU acceleration is still a challenging issue. This is why SDS could in theory save computation, but in practice that doesn't happen (see previous studies of Dynamic Sparse Training~\cite{bellec2017deep,mocanu2018scalable,dettmers2019sparse,liu2021sparse,evci2020rigging,mostafa2019parameter,jayakumar2020top,chen2021chasing,liu2021selfish} for related discussion). Thus, only small speed change (4ms) is observed with SDS in Tab.\!~{\color{red}6}. Our speed up mainly comes from CWS. CWS can significantly reduce the model size at the channel level, leading to a primary acceleration of \textbf{84} ms (177ms$\rightarrow$93ms). Moreover, for a similar reason, we present the theoretical inference FLOPs of the models under different settings.

\noindent\textbf{CWS.}~Our CWS is of great importance for building 3D large kernels.$_{\!}$ Previous 2D large sparse kernels like \cite{liu2022more} follows a regime of ``\textit{use sparse groups, expand more}''; they attain better performance with enlarged network width, inevitably causing increased model size. This becomes more pronounced in 3D. For a naive dense kernel, lifting ($64$-$d$, $9\!\times\!9\!\times\!9$) to ($128$-$d$, $9\!\times\!9\!\times\!9$) will cause a significant (somewhat unacceptable) increase of parameter number: 47.94M $\rightarrow$ 191.69M. As shown in Tab.\!~{\color{red}7} of the main paper, LSK3DNet (with CWS) successfully achieves better performance while using fewer parameters.

\noindent\textbf{Kernel Size.}~Due to the increased dimensionality, enlarging 3D kernel size brings much heavier computational burden, compared with 2D kernel.$_{\!}$ As shown in Tab.\!~{\color{red}5} of the main paper, comparing Dense ($7\!\times\!7\!\times\!7$, $1\!\times\!D$) and Dense ($9\!\times\!9\!\times\!9$, $1\!\times\!D$), computational burden$_{\!}$ has$_{\!}$ \textbf{5}$\times$$_{\!}$ rise:$_{\!}$ 381.2G $\!\rightarrow\!$ 1916.3G.$_{\!}$ Though$_{\!}$ large$_{\!}$ kernel gains larger receptive fields, it brings risk of overparameterization and overfitting, and is hard to train. \cite{ma2022hyper,choy20194d,chen2022scaling,liu2022more} encounter a similar issue, \eg, increasing $3\!\times\!3\!\times\!3$ to $7\!\times\!7\!\times\!7$ leads to inferior results~\cite{chen2022scaling}, $31\!\times\!31$ performs worse than $7\!\times\!7$~\cite{liu2022more}.

\noindent\textbf{Convolutional Schemes.}~The results of MinkowskiNet-14 and MinkowskiNet-34 are directly taken from LargeKernel3D~\cite{chen2022scaling}.$_{\!}$ MinkowskiNet-14 shows stagnation when attempting to expand the convolution kernel to $9\!\times\!9\!\times\!9$. As shown in Tab.\!~\ref{tab:groupconvolution}, we use Modified SPVCNN as the baseline and conduct experiments on ScanNet v2~\cite{dai2017scannet} and SemanticKITTI~\cite{behley2019semantickitti}. Our findings indicate that \textit{Spatial-wise Group Convolution} is more suitable for 3D large kernel designs, aligning with the results observed in LargeKernel3D~\cite{chen2022scaling}. Therefore, we carry out dynamic sparse training within the spatial-wise groups.

\section{More Qualitative and Quantitative Results}
\label{appendix_3}
\noindent\textbf{Quantitative Results on NuScenes.}~In Tab.\!~\ref{tab:nus_seg_valid}, we present the results of semantic segmentation on the nuScenes validation set~\cite{caesar2020nuscenes}.$_{\!}$ Our approach consistently surpasses others by a significant margin, establishing itself as the state-of-the-art (SOTA) performer on this benchmark.$_{\!}$ What's particularly intriguing is that our method relies solely on LiDAR data, yet it outperforms multi-modal techniques~\cite{genova2021learning,yan20222dpass}, which incorporate supplementary 2D information.

\noindent\textbf{Concrete Results on SemanticKITTI Multi-scan Test.}~The full results of our LSK3DNet on the SemanticKITTI~\cite{behley2019semantickitti} multi-scan \texttt{test} challenge are shown in Tab.~\ref{table:4DSegmentationtest1appendix} and \ref{table:4DSegmentationtest1appendix_3}.$_{\!}$ Our method achieves a state-of-the-art performance on both mIoU and Acc metrics, surpassing other methods in 12 out of the 25 categories. Unlike the previous method 2DPASS, which relies on 2D images as an auxiliary input, our algorithm is purely based on point cloud data, which is a valuable advantage.

\begin{figure}[t]
  \centering
      \includegraphics[width=1.01 \linewidth]{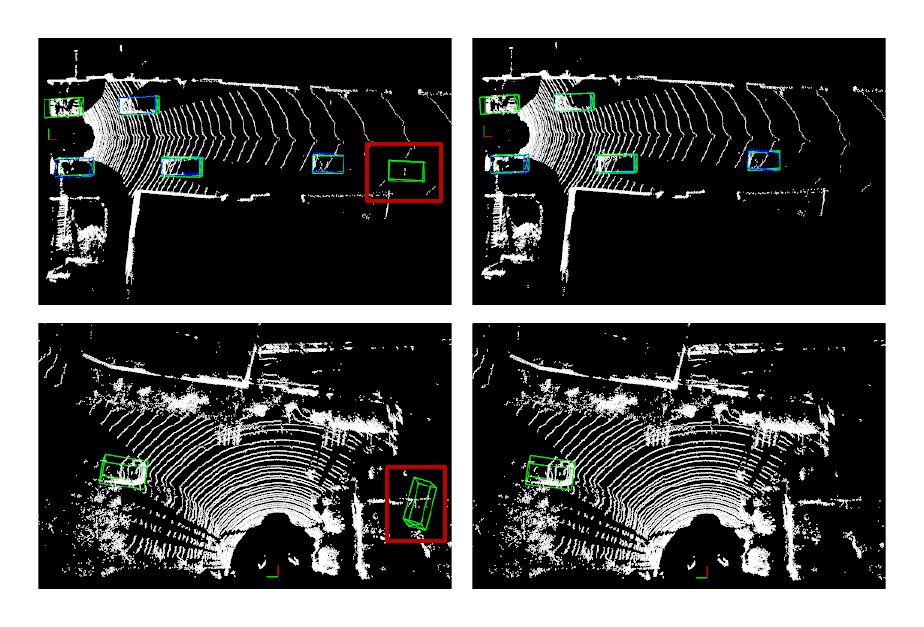}
      \put(-204,-6){\scalebox{.80}{Voxel R-CNN}}
      \put(-75,-6){\scalebox{.80}{LSK3DNet}}
\vspace{-6pt}
\captionsetup{font=small}
\caption{\small \textbf{Qualitative detection results} of LSK3DNet and Voxel R-CNN on the KITTI \textit{val} split (Sec.\!~{\color{red}4.3}). Results in the red box are false positives.}
\label{fig:kittival}
\end{figure}

\noindent\textbf{Qualitative Detection Results for KITTI.}~We compare the visualization results between LSK3DNet and Voxel R-CNN for the car category on the KITTI \textit{val} split~\cite{geiger2013vision}, as illustrated in Fig.\!~\ref{fig:kittival}.$_{\!}$ The blue-bordered boxes depict ground truth bounding boxes, while the green-bordered boxes represent predicted bounding boxes. In comparison to Voxel R-CNN, LSK3DNet demonstrates more accurate prediction results.

\begin{figure*}[h!]
  \centering
      \includegraphics[width=0.98 \linewidth]{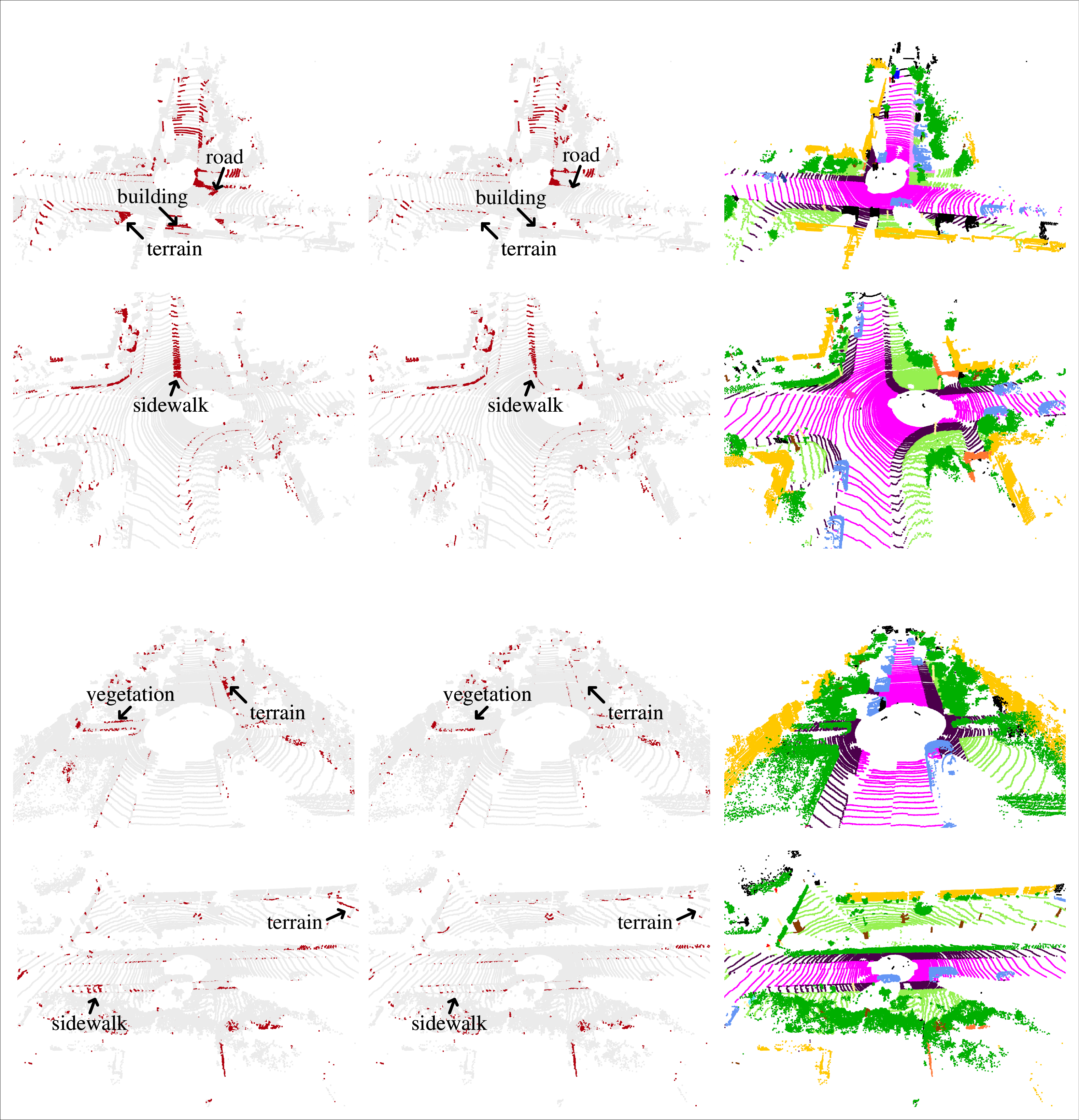}
      \put(-430.5,-6) {\scalebox{.80}{Error by Baseline}}
      \put(-255.5,-6){\scalebox{.80}{Error by LSK3DNet}}
      \put(-100.5,-6){\scalebox{.80}{Ground Truth}}

\caption{Error maps of Baseline and Ours on SemanticKITTI~\cite{behley2019semantickitti} single-scan \texttt{val}$_{\!}$ (Sec.\!~{\color{red}4.2}).}
\label{fig:vresults1appendix}
\end{figure*}

\begin{figure*}[h!]
  \centering
      \includegraphics[width=0.98 \linewidth]{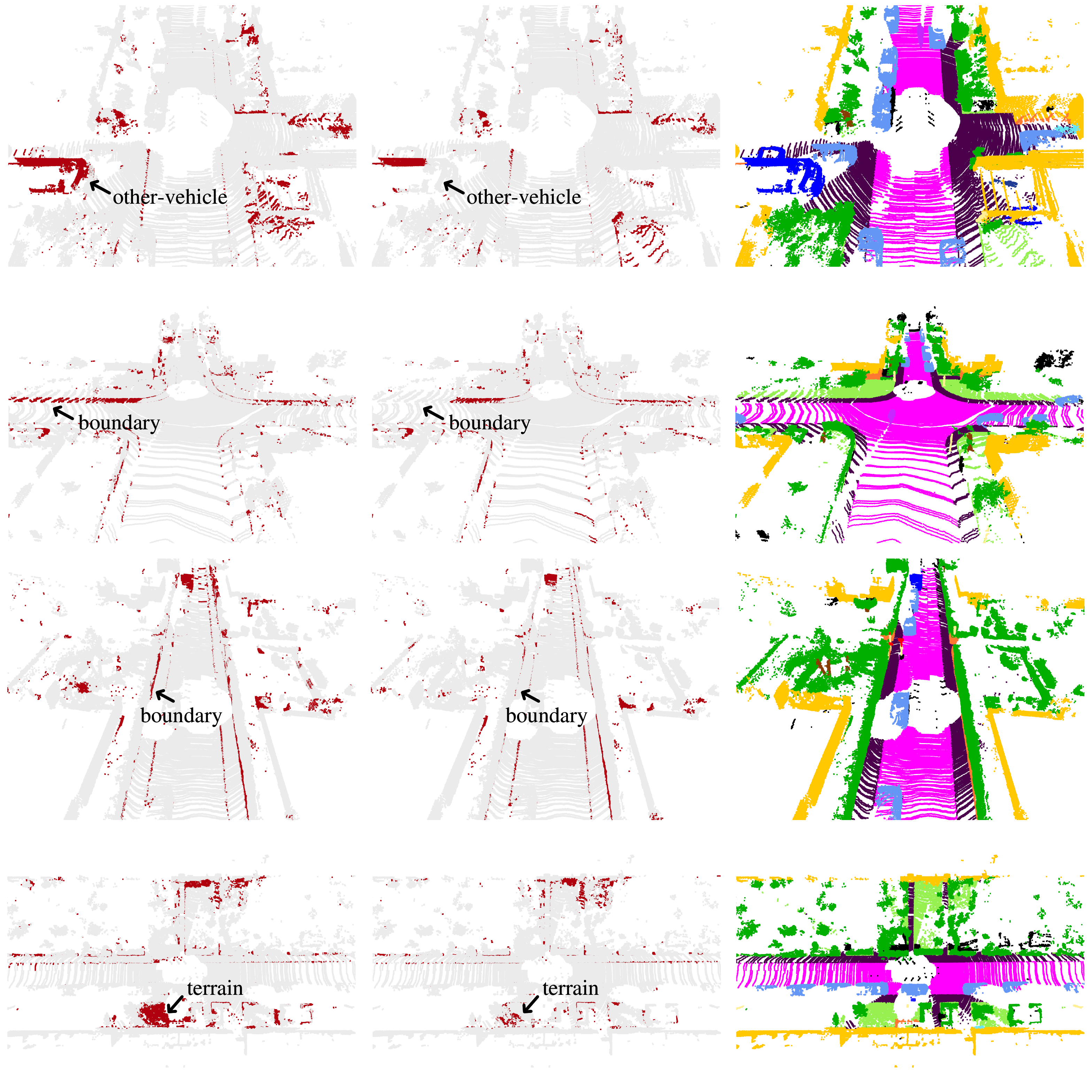}
      \put(-430.5,-6) {\scalebox{.80}{Error by Baseline}}
      \put(-255.5,-6){\scalebox{.80}{Error by LSK3DNet}}
      \put(-100.5,-6){\scalebox{.80}{Ground Truth}}

\caption{Error maps of Baseline and Ours on SemanticKITTI~\cite{behley2019semantickitti} multi-scan \texttt{val}$_{\!}$ (Sec.\!~{\color{red}4.2}).}
\label{fig:vresults3appendix}
\end{figure*}

\noindent\textbf{Qualitative Results for SemanticKITTI.}~We provide visual examples of our algorithm on two challenges for 3D semantic segmentation: SemanticKITTI~\cite{behley2019semantickitti} single-scan \texttt{val} challenge and SemanticKITTI~\cite{behley2019semantickitti} multi-scan \texttt{test} challenge.$_{\!}$ The corresponding figures are Fig.~\!\ref{fig:vresults1appendix} and Fig.~\!\ref{fig:vresults3appendix}, respectively. Moreover, we display the predicted labels of three successive frames in one single frame in Fig.~\!\ref{fig:vresults3appendix}. It can be seen that our LSK3DNet produces more accurate predictions than the baseline (Modified SPVCNN~\cite{yan20222dpass}). By a larger kernel size, our LSK3DNet expands the receptive field of submanifold convolution and enhances the flow of discrete spatial information. This results in better capturing the object boundaries and distinguishing between different semantic classes. As shown, our LSK3DNet can segment the ground classes and natural objects more effectively.

\section{Limitation and Societal Impact}
\label{appendix_4}

\noindent\textbf{Limitation.}~Dense matrix multiplications on graphics processing units (GPUs) are the foundation of the current state-of-the-art deep learning methods.$_{\!}$ However, sparse matrix multiplications,$_{\!}$ which are crucial for Dynamic Sparse Training operations, cannot be efficiently performed~\cite{lebedev2016fast,changpinyo2017power,mocanu2018scalable}.$_{\!}$ Hence, the FLOPs for inference that we present in the main paper are based on theoretical calculations. This creates a need for optimized hardware that can handle such operations.$_{\!}$ However, there is a good news that sparsity support is becoming a more common feature in hardware design for many companies and researchers~\cite{liu2021sparse,gale2020sparse,choquette2021nvidia}.$_{\!}$ Our work aims to inspire more progress in this direction, especially for 3D tasks.

\noindent\textbf{Societal Impact.}~Self-driving cars need to efficiently and accurately understand 3D scenes to ensure safe driving. Since passenger safety is the top priority for autonomous vehicles, 3D perception models are required to achieve both high accuracy and low latency. Autonomous vehicles have limited hardware resources due to the physical limitations of size and heat dissipation. Therefore, it is important to design 3D neural networks that are both efficient and effective for constrained computing resources. We use the method of Spatial Sparse Group to further expand the 3D convolution kernel size, which overcomes the performance drop of previous 3D large kernel methods. Moreover, the model is very efficient due to dynamic sparse and \textit{Channel-wise Weight Selection}. 

\noindent\textbf{License.}~We study 3D semantic segmentation and object detection on four famous datasets.$_{\!}$ We use the SemanticKITTI dataset under the permission of its creators and authors by registering at \url{https://codalab.lisn.upsaclay.fr/competitions/6280}.$_{\!}$ Scannet v2 is released under the ScanNet Terms of Use (\url{https://kaldir.vc.in.tum.de/scannet/ScanNet_TOS.pdf}).$_{\!}$ KITTI is published under the Creative Commons Attribution-NonCommercial-ShareAlike 3.0 License.$_{\!}$ nuScenes is available for non-commercial use subject to the Terms of Use (See \url{https://www.nuscenes.org/terms-of-use}).

\noindent\textbf{Computing Infrastructure.}~The hardware configuration consists of an Intel(R) Xeon(R) Gold 6132 CPU @ 2.60GHz and 18GB of memory.$_{\!}$ The operating system is Ubuntu 18.04. All the experiments use Tesla V100 GPUs with 32GB of VRAM. Moreover, our method is executed on top of the PyTorch framework.

\newpage

\begin{table*}[t]
\small
\caption{\small \textbf{Semantic segmentation results} on NuScenes \texttt{validation} set~\cite{caesar2020nuscenes}. Regarding input data format, P denotes points, V represents voxelizations, R signifies range images, 2D3DNet and 2DPASS incorporate additional 2D data. mIoU (\%) and IoUs (\%) are reported.}
\label{tab:nus_seg_valid}
\centering
\setlength\tabcolsep{4.8pt}
\renewcommand\arraystretch{1.0}
\resizebox{\linewidth}{!}{
\begin{tabular}{r|c|c|cccccccccccccccc}
\hline
Method& \rotatebox{90}{Input} & \rotatebox{90}{\bf{mIoU}}& \rotatebox{90}{barrier}& \rotatebox{90}{bicycle}& \rotatebox{90}{bus}& \rotatebox{90}{car}& \rotatebox{90}{construction~}& \rotatebox{90}{motorcycle}& \rotatebox{90}{pedestrian}& \rotatebox{90}{traffic cone}& \rotatebox{90}{trailer}& \rotatebox{90}{truck}& \rotatebox{90}{driveable}& \rotatebox{90}{other flat}& \rotatebox{90}{sidewalk}& \rotatebox{90}{terrain}& \rotatebox{90}{manmade}& \rotatebox{90}{vegetation}\\
\hline
\hline
(AF)$^2$-S3Net\!~\cite{cheng20212} & V &62.2 &60.3  &12.6  &82.3 & 80.0  &  20.1 & 62.0 & 59.0 & 49.0 &  42.2  & 67.4 & 94.2 &  68.0  &  64.1  &  68.6  &  82.9  & 82.4  \\
RangeNet++\!~\cite{milioto2019rangenet++} & R & 65.5 & 66.0 & 21.3 & 77.2 & 80.9 & 30.2 & 66.8 & 69.6 & 52.1 & 54.2 & 72.3 & 94.1 & 66.6 & 63.5 & 70.1 & 83.1 & 79.8 \\
PolarNet\!~\cite{zhang2020polarnet} & R & 71.0 & 74.7 & 28.2 & 85.3 & 90.9 & 35.1 & 77.5 & 71.3 & 58.8 & 57.4 & 76.1 & 96.5 & 71.1 & 74.7 & 74.0 & 87.3 & 85.7\\
Salsanext\!~\cite{cortinhal2020salsanext} & R & 72.2 & 74.8 & 34.1 & 85.9 & 88.4 & 42.2 & 72.4 & 72.2 & 63.1 & 61.3 & 76.5 & 96.0 & 70.8 & 71.2 & 71.5 & 86.7 & 84.4 \\
AMVNet\!~\cite{liong2020amvnet} & P & 76.1 & 79.8 & 32.4 & 82.2 & 86.4 & 62.5 & 81.9 & 75.3 & 72.3 & 83.5 & 65.1 & 97.4 & 67.0 & 78.8 & 74.6 & 90.8 &87.9 \\
Cylinder3D\!~\cite{zhu2021cylindrical} & V & 76.1 & 76.4 & 40.3 & 91.2 & 93.8 & 51.3 & 78.0 & 78.9 & 64.9 & 62.1 & 84.4 & 96.8 & 71.6 & 76.4 & 75.4 & 90.5 & 87.4 \\
RPVNet\!~\cite{xu2021rpvnet} & RPV & 77.6 & 78.2 & 43.4 & 92.7 & 93.2 & 49.0 & 85.7 & 80.5 & 66.0 & 66.9 & 84.0 & 96.9 & 73.5 & 75.9 & 76.0 & 90.6 & 88.9 \\
PVKD\!~\cite{hou2022point} & V &76.0 & 76.2 & 40.0& 90.2& 94.0 &50.9& 77.4& 78.8 &64.7 &62.0& 84.1 &96.6& 71.4 &76.4 &76.3 & 90.3 & 86.9\\
2D3DNet\!~\cite{genova2021learning} & V & 79.0 & 78.3 & 55.1 & 95.4 & 87.7 & 59.4 & 79.3 & 80.7 & 70.2 & 68.2 & 86.6 & 96.1 & 74.9 & 75.7 & 75.1 & 91.4 & 89.9 \\
2DPASS\!~\cite{yan20222dpass}  & PV& 79.4 & 78.8 & 49.6 & 95.6 & 93.6 & 60.0 & 84.1 & 82.2 & 67.5 & 72.6 & 88.1 & 96.8 & 72.8 & 76.2 & 76.5 & 89.4 & 87.2\\
SphereFormer\!~\cite{lai2023spherical} & V & 79.5 &78.7 &46.7 &95.2& 93.7 &54.0& 88.9& 81.1& 68.0& 74.2& 86.2& 97.2& 74.3 & 76.3 &75.8 & 91.4 & 89.7 \\
\hline
LSK3DNet & PV &\bf{80.1}& 80.0& 52.5& 94.5 & 91.8 & 58.8&85.8&84.4&71.2&73.8&88.3&96.9&74.8&75.9&75.9&89.3&87.5\\
\hline
\end{tabular}}
\vspace{1.5pt}
\end{table*}

\begin{table*}[h!]
\begin{center}
\caption{\small $_{\!}$\textbf{Quantitative$_{\!}$ results$_{\!}$} on$_{\!}$ SemanticKITTI$_{\!}$~\cite{behley2019semantickitti}$_{\!}$ multi-scan$_{\!}$ challenge$_{\!}$ \texttt{test}$_{\!}$  (Sec.\!~{\color{red}4.2}) - Part I. mIoU (\%) and IoUs (\%) are reported.}
\label{table:4DSegmentationtest1appendix}
\setlength\tabcolsep{7.5pt}
\renewcommand\arraystretch{1.18}
\resizebox{\linewidth}{!}{
\begin{tabular}{r|cc|ccccccccccccc}
\hline
Method&mIoU&Acc&\rotatebox{90}{road}&\rotatebox{90}{sidewalk}&\rotatebox{90}{parking}&\rotatebox{90}{other-ground}&\rotatebox{90}{building}&\rotatebox{90}{car}&\rotatebox{90}{moving car}&\rotatebox{90}{truck}&\rotatebox{90}{moving truck}&\rotatebox{90}{bicycle}&\rotatebox{90}{motorcycle}&\rotatebox{90}{other-vehicle}&\rotatebox{90}{moving other-vehicle}\\
\hline
\hline
TangentConv\!~\cite{tatarchenko2018tangent}&34.1&-&83.9&64.0&38.3&15.3&85.8&84.9&40.3&21.1&1.1&2.0&18.2&18.5&6.4\\
DarkNet53\!~\cite{behley2019semantickitti}&41.6&-&91.6&75.3&64.9&27.5&85.2&84.1&61.5&20.0&14.1&30.4&32.9&20.7&15.2\\
TemporalLidarSeg\!~\cite{duerr2020lidar}&47.0&89.6&91.8&75.8&59.6&23.2&89.8&92.1&68.2&39.2&2.1&47.7&40.9&35.0&12.4\\
SpSeqnet\!~\cite{shi2020spsequencenet}&43.1&-&90.1&73.9&57.6&27.1&91.2&88.5&53.2&29.2& 41.2&24.0&26.2&22.7&26.2\\
KPConv\!~\cite{thomas2019kpconv}&51.2&89.3&86.5&70.5&58.4&26.7&90.8&93.7&69.4&42.5&5.8&44.9&47.2&38.6&4.7\\
Cylinder3D\!~\cite{zhu2021cylindrical}&52.5&91.0&90.7&74.5&65.0&32.3&92.6&94.6&74.9&41.3&0.0&67.6&63.8&38.8&0.1\\
(AF)$^2$-S3Net\!~\cite{cheng20212} & 56.9&88.1 &91.3& 72.5& 68.8& 53.5& 87.9& 91.8& 65.3& 15.7&5.6& 65.4& 86.8& 27.5&3.9\\ 
PV-KD\!~\cite{hou2022point} &58.2&91.9& 92.4 &77.4& 69.9&31.5&92.7&96.2&84.3&50.0&20.9&64.9&64.8&46.4&19.0\\
2DPASS\!~\cite{yan20222dpass} & 62.4&91.4 &89.7&74.7&67.4&40.0&93.6&96.2&82.1&48.2&16.1&63.6&63.7&52.7&3.8\\ 
\hline
LSK3DNet &\bf{63.4}& \bf{92.2}& 92.1& 79.0& 67.4 & 41.0&92.9& 96.4 & 84.4 & 58.1&7.2& 71.2&73.9&61.8&40.9\\
\hline
\end{tabular}}
\end{center}
\vspace{-7.5pt}
\end{table*}

\begin{table*}[h!]
\begin{center}
\caption{\small $_{\!}$\textbf{Quantitative$_{\!}$ results$_{\!}$} on$_{\!}$ SemanticKITTI$_{\!}$~\cite{behley2019semantickitti}$_{\!}$ multi-scan$_{\!}$ challenge$_{\!}$ \texttt{test}$_{\!}$  (Sec.\!~{\color{red}4.2}) - Part II. mIoU (\%) and IoUs (\%) are reported.
}
\label{table:4DSegmentationtest1appendix_3}
\setlength\tabcolsep{8.5pt}
\renewcommand\arraystretch{1.18}
\resizebox{\linewidth}{!}{
\begin{tabular}{r|cc|cccccccccccc}
\hline
Method&mIoU&Acc&\rotatebox{90}{vegetation}&\rotatebox{90}{trunk}&\rotatebox{90}{terrain}&\rotatebox{90}{person}&\rotatebox{90}{moving person}&\rotatebox{90}{bicyclist}&\rotatebox{90}{moving bicyclist}&\rotatebox{90}{motorcyclist}&\rotatebox{90}{moving motorcyclist}&\rotatebox{90}{fence}&\rotatebox{90}{pole}&\rotatebox{90}{traffic-sign}\\
\hline
\hline
TangentConv\!~\cite{tatarchenko2018tangent}&34.1&-&79.5& 43.2& 56.7&1.6&1.9&0.0&30.1&0.0&42.2&49.1&36.4&31.2\\
DarkNet53\!~\cite{behley2019semantickitti}&41.6&-&78.4& 50.7& 64.8&7.5&0.2&0.0&28.9&0.0&37.8&56.5&38.1&53.3\\
TemporalLidarSeg\!~\cite{duerr2020lidar}&47.0&89.6&82.3&62.5&64.7&14.4&40.4&0.0&42.8&0.0&12.9&63.8&52.6&60.4\\
SpSeqnet\!~\cite{shi2020spsequencenet}&43.1&-&84.0& 66.0&65.7&6.3&36.2&0.0&2.3&0.0&0.1&66.8&50.8&48.7 \\
KPConv\!~\cite{thomas2019kpconv}&51.2&89.3&84.6&70.3&66.0&21.6&67.5&0.0&67.4&0.0&47.2&64.5&57.0&53.9\\
Cylinder3D\!~\cite{zhu2021cylindrical}&52.5&91.0&85.8&72.0&68.9&12.5&65.7&1.7&68.3&0.2&11.9&66.0&63.1&61.4\\
(AF)$^2$-S3Net\!~\cite{cheng20212} & 56.9&88.1 &75.1&64.6& 57.4& 16.4& 67.6&15.1&66.4&67.1&59.6&63.2&62.6&71.0\\
PV-KD\!~\cite{hou2022point} &58.2&91.9&86.4&74.1&70.2&16.6&68.5&0.0&69.2&2.0&50.5&70.3&66.9&70.6\\
2DPASS\!~\cite{yan20222dpass} & 62.4&91.4&86.2&73.9&71.0&35.4&80.3&7.9&71.2&62.0&73.1&72.9&65.0&70.5\\
\hline
LSK3DNet &\bf{63.4}&\bf{92.2}& 86.9&74.2&72.6&29.3&77.4&0.1&69.9&22.8&72.1&72.3&66.7&74.3\\
\hline
\end{tabular}}
\end{center}
\vspace{-21pt}
\end{table*}

\newpage

\newpage

\clearpage